\documentclass[journal]{IEEEtran}
\IEEEoverridecommandlockouts
\usepackage{cite}
\usepackage{amsmath,amssymb,amsfonts}
\usepackage{graphicx}
\usepackage{textcomp}
\usepackage{xcolor}
\usepackage{multicol}
\usepackage{mathabx}
\usepackage{soul}
\usepackage{hyperref}
\usepackage{enumerate}
\usepackage{subcaption}
\usepackage{comment}

\usepackage{bbm}
\usepackage{float}
\usepackage{algorithmic}
\usepackage[linesnumbered, ruled, vlined, noend]{algorithm2e}
\usepackage{tcolorbox}
\usepackage[normalem]{ulem}
\tcbuselibrary{theorems}

\newtcbtheorem{mytheo}{Comment}%
{colback=gray!5,colframe=gray!35!black,fonttitle=\bfseries}{th}

\def\BibTeX{{\rm B\kern-.05em{\sc i\kern-.025em b}\kern-.08em
    T\kern-.1667em\lower.7ex\hbox{E}\kern-.125emX}}

\begin{document}

\title{Optimizing Energy and Data Collection in UAV-aided IoT Networks using Attention-based Multi-Objective Reinforcement Learning
}

\author{
Babacar Toure, Dimitrios Tsilimantos, Omid Esrafilian, and Marios Kountouris
\thanks{Babacar Toure (Corresponding author) is with the Advanced Wireless Technology Lab, Paris Research Center, Huawei Technologies France and also with EURECOM, Sophia Antipolis, France (email:babacar.toure1@huawei.com).}
\thanks{Dimitrios Tsilimantos is with the Advanced Wireless Technology Lab, Paris Research Center, Huawei Technologies France (email: dimitrios.tsilimantos@huawei.com).}
\thanks{Omid Esrafilian is with the Communication Systems Department, EURECOM, Sophia Antipolis, France (email: omid.esrafilian@eurecom.fr).}
\thanks{Marios Kountouris is with the Dept. of Computer Science and Artificial Intelligence, University of Granada, Spain and also with EURECOM, Sophia Antipolis, France (mariosk@ugr.es).}
}

\newcommand{\myitem}[1]{\vspace{0.25\baselineskip}\noindent\textbf{#1}}
\newcommand{\Hquad}{\hspace{0.5em}}

\maketitle

\begin{abstract}

Due to their adaptability and mobility, Unmanned Aerial Vehicles (UAVs) are becoming increasingly essential for wireless network services,  particularly for data harvesting tasks. In this context, Artificial Intelligence (AI)-based approaches have gained significant attention for addressing UAV path planning tasks in large and complex environments, bridging the gap with real-world deployments. However, many existing algorithms suffer from limited training data, which hampers their performance in highly dynamic environments. Moreover, they often overlook the inherently multi-objective nature of the task, treating it in an overly simplistic manner. To address these limitations, we propose an attention-based Multi-Objective Reinforcement Learning (MORL) architecture that explicitly handles the trade-off between data collection and energy consumption in urban environments, even without prior knowledge of wireless channel conditions. Our method develops a single model capable of adapting to varying trade-off preferences and dynamic scenario parameters without the need for fine-tuning or retraining. Extensive simulations show that our approach achieves substantial improvements in performance, model compactness, sample efficiency, and most importantly, generalization to previously unseen scenarios, outperforming existing RL solutions.

\end{abstract}


\section{Introduction}
\label{sec:introduction}
\bstctlcite{IEEEbibChanges:BSTcontrol}
\IEEEPARstart{U}{nmanned} Aerial Vehicles (UAVs) are expected to play a key role in next-generation wireless communication networks, due to their versatility and ease of deployment. UAVs support a wide range of Internet of Things (IoT) applications, including wireless coverage, data relaying, data collection, edge computing, and disaster response \cite{10113154}. In particular, for data collection, UAVs offer an efficient and scalable solution to address the growing volume of IoT devices in agricultural, smart city, and industrial environments. Compared to ground-based infrastructure alone, UAVs provide mobility, rapid deployment, and operational flexibility. They can adapt their flight paths in real time based on network conditions, access remote or hazardous areas with poor connectivity, and help reduce both the cost and latency of data collection \cite{uav_data_collection_survey}.

However, UAV-based data collection tasks often lead to NP-hard and non-convex optimization problems \cite{hu2020distributedmultiagentmetalearning, 8847391, 8736350}, mainly due to the combinatorial complexity of trajectory planning, the absence of reliable communication models, and the presence of multiple, often conflicting, objectives. This complexity is further exacerbated by the dynamic and uncertain nature of the environments in which UAVs operate, characterized by high-dimensional state and action spaces. In this context, Deep Reinforcement Learning (DRL) has emerged as a powerful tool for learning effective navigation policies by interacting with the environment, without relying on predefined models or assumptions. Moreover, like many real-world applications, UAV data collection inherently involves balancing multiple objectives, such as minimizing energy consumption, maximizing data yield, and optimizing data freshness \cite{uav_data_collection_survey}. In this paper, our goal is to address these challenges by leveraging DRL algorithms that are not only scalable, sample-efficient, and stable but also capable of handling multiple objectives and producing interpretable policies.

\subsection{Related Work}
Path planning for UAV-enabled IoT applications has become a highly active research topic in recent years. According to the comprehensive survey in \cite{uav_data_collection_survey}, existing data collection strategies fall into three broad categories: (i) conventional heuristics, (ii) optimization-based methods, and (iii) AI-driven techniques.

Conventional path-planning methods typically leverage graph-based algorithms such as Dijkstra, A*, Voronoi diagrams, and related techniques. These approaches model the navigation environment as a graph, where nodes represent discrete waypoints and edges denote potential flight paths. Task-specific metrics, such as the volume of collected data, the data freshness, and the energy consumption, are commonly used to guide heuristic decisions within these algorithms. For example, \cite{Tao2018PathPA} introduces an A* path planning algorithm based on an efficacy function that accounts for data value, energy usage, mission duration, and risk, aiming to address UAV energy constraints. Similarly, \cite{Alejo2015EfficientTP} employs a Rapidly-exploring Random Tree (RRT)-based planner to facilitate data collection from a Wireless Sensor Network (WSN), demonstrating its effectiveness in both simulation and real-world settings. Although these methods perform well in small- to medium-scale scenarios, they generally suffer from limited scalability and reduced robustness in dynamic environments. In particular, environmental changes often require reconstruction of the underlying graph, a process that can be computationally intensive and inefficient in real-time operations.

Optimization algorithms have also been extensively applied to tackle data collection challenges in IoT systems. In \cite{liu2018ageoptimaltrajectoryplanninguavassisted}, an optimal Age of Information (AoI) trajectory is designed for a UAV that collects data from ground sensor nodes in a wireless sensor network (WSN). The authors employ dynamic programming and genetic algorithms to derive two distinct trajectories: one that minimizes the average AoI across all sensor nodes, and another that minimizes the maximum AoI. Similarly, \cite{10334482} explores a related scenario, but focuses on the trade-off between AoI and energy consumption. The resulting non-convex optimization problem is decomposed into three sub-problems, which are solved using Successive Convex Approximation and Lagrangian duality techniques. Simulation results demonstrate that simultaneously minimizing both AoI and energy consumption is generally infeasible, highlighting the need to consider such problems from a multi-objective perspective, where multiple Pareto-optimal solutions may exist. Optimization-based approaches are valued for their convergence guarantees and mathematical rigor. However, they often lack adaptability in real-time scenarios and tend to generalize poorly to changing environmental conditions or problem configurations. In most cases, solutions must be re-tuned or re-optimized when applied to new instances, limiting their scalability and practical deployment.

Recently, RL algorithms have shown promising performance in UAV path-planning tasks, addressing several limitations of traditional methods. These algorithms learn adaptive navigation policies by interacting with the environment, without requiring an explicit model of its dynamics. A notable early contribution is presented in \cite{9162896}, which tackles the problem of data freshness under energy constraints using a Deep Q-Network (DQN). The use of DQN specifically mitigates the curse of dimensionality by employing Deep Neural Networks (DNNs) as function approximators, enabling near-optimal decision-making in large state spaces and facilitating real-time execution with strong generalization capabilities. The proposed approach outperforms conventional heuristic-based methods.
In a similar vein, \cite{9174950} addresses the joint optimization of UAV flight cruise control and scheduling, aiming to minimize data loss on IoT devices caused by buffer overflows. Meanwhile, \cite{9504602} employs the Twin-Delayed Deep Deterministic Policy Gradient (TD3) algorithm to maximize throughput during UAV navigation, reporting substantial gains over non-learning baselines in realistic simulation scenarios. Furthermore, \cite{9209109} explores a dual-purpose use of DRL for simultaneous UAV-based wireless energy transfer and data collection.
Despite their advantages, DRL models often exhibit limited generalization to new environments or varying problem parameters. This shortcoming is largely due to the reliance on fixed network architectures and rigid state representations. As shown in \cite{9437338}, multi-UAV frameworks in particular require retraining when key environmental variables change, which is computationally intensive given the large number of interactions typically required for RL convergence. To overcome this limitation, the authors of \cite{9437338} propose an extended approach in subsequent works \cite{9437338, 9322234, Theile_2021}, introducing convolutional neural network (CNN) architectures in place of standard feedforward networks. They also propose a novel spatial map representation that integrates a compressed global view with a dynamically updated local map. This hybrid representation significantly improves generalization across diverse environments and scales effectively to larger operational areas.

\begin{figure}[h]
  \centering
  \begin{subfigure}{.16\textwidth}
    \centering
    \includegraphics[width=\textwidth]{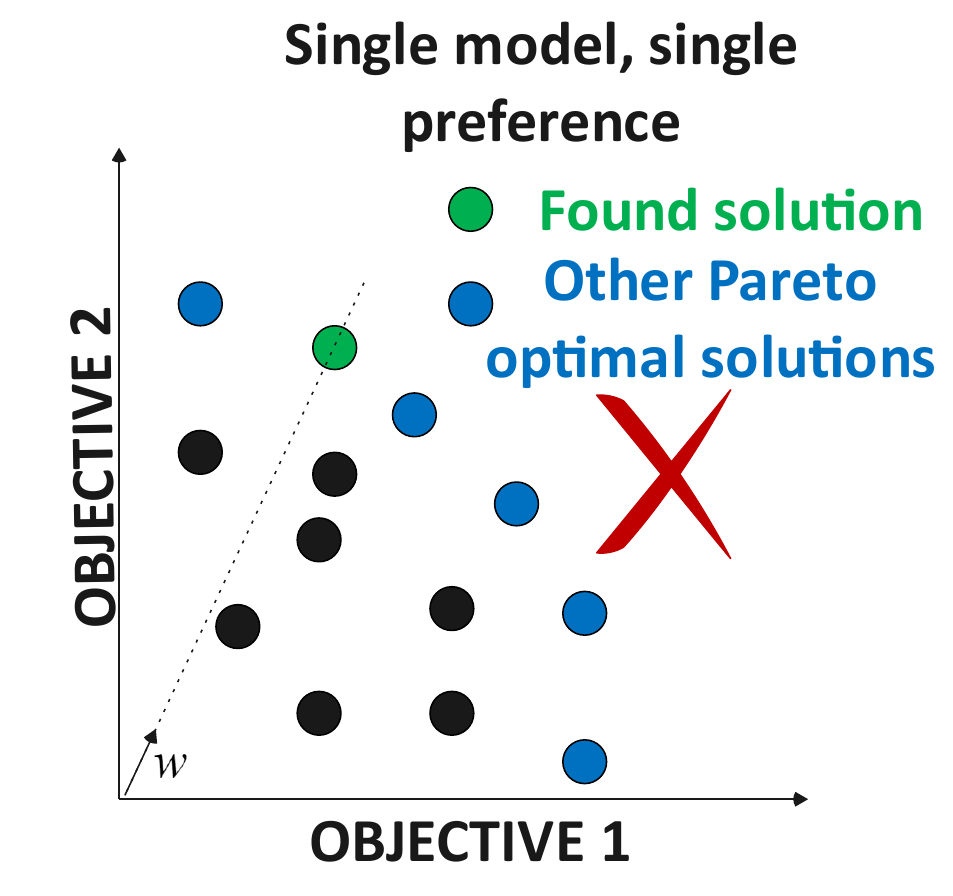}
    \caption{Naive}
    \label{fig:target_front_1}
  \end{subfigure}%
  \hfill
  \begin{subfigure}{.16\textwidth}
    \centering
    \includegraphics[width=\textwidth]{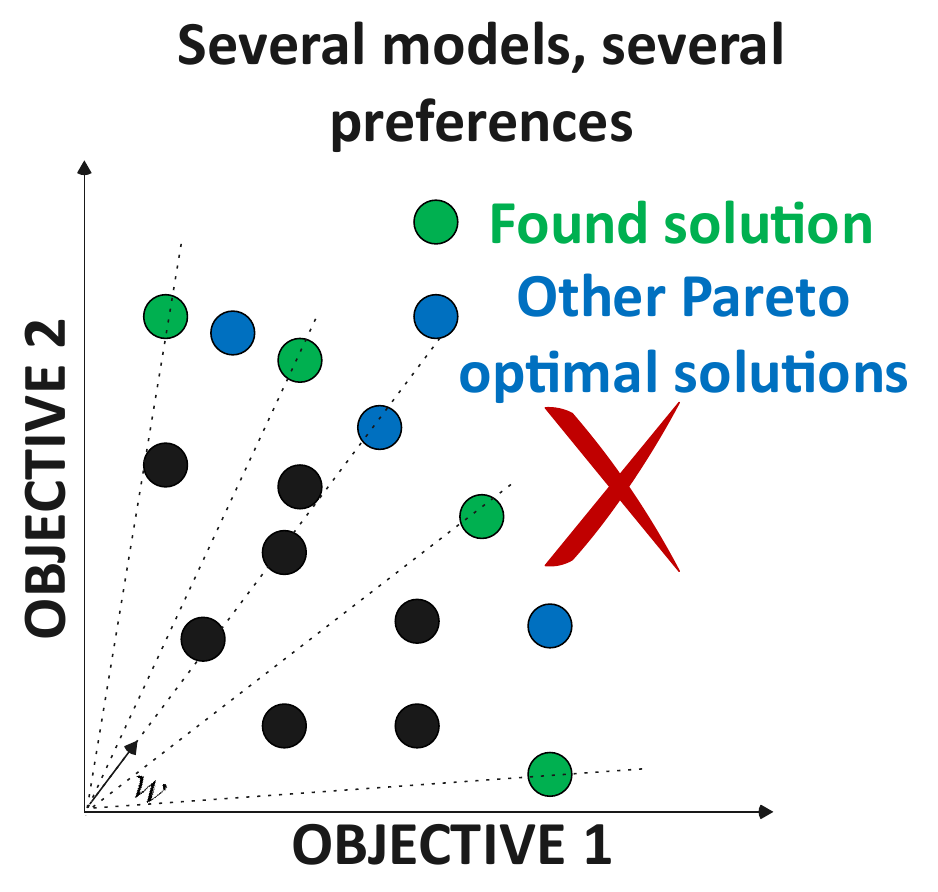}
    \caption{Costly}
    \label{fig:target_front_2}
  \end{subfigure}%
  \hfill
  \begin{subfigure}{.16\textwidth}
    \centering
    \includegraphics[width=\textwidth]{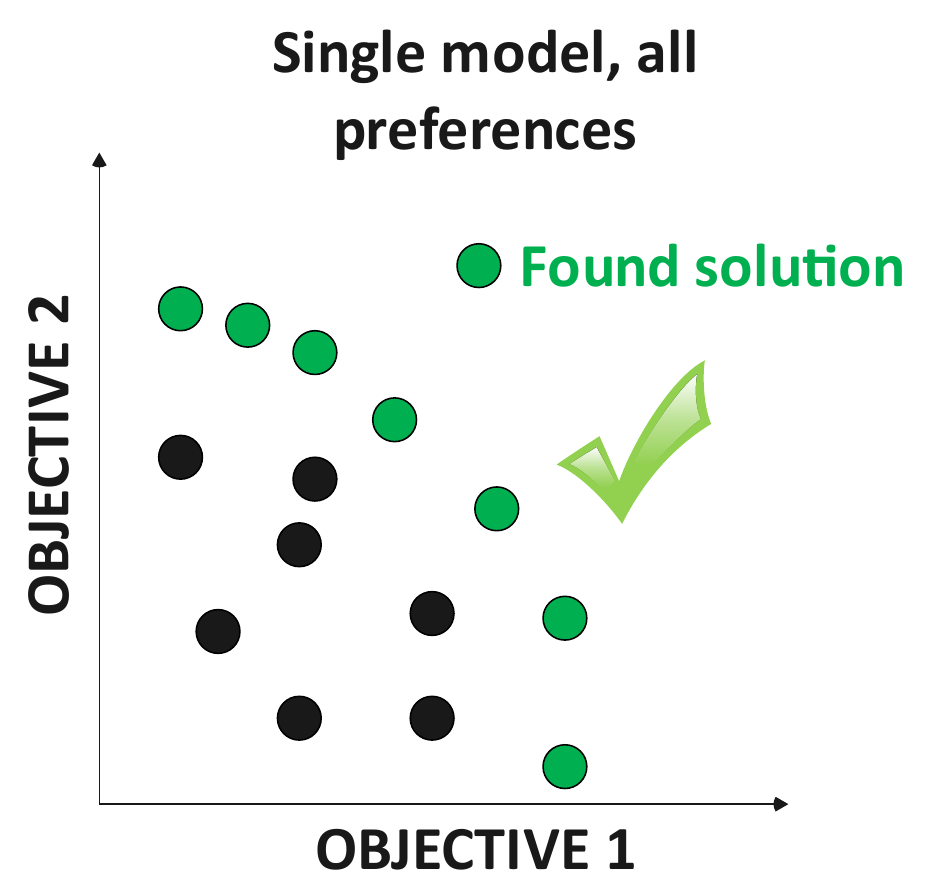}
    \caption{Desired }
    \label{fig:target_front_3}
  \end{subfigure}%
  \hfill
  \caption{Different approaches in multi-objective learning.}
  \label{fig:target_front}
\end{figure}

An often overlooked aspect of UAV-based data collection is the inherently multi-objective nature of the task. Many existing studies, including those cited above, address problems involving conflicting objectives, such as maximizing data collection, minimizing energy consumption, and optimizing data freshness. In the worst-case scenario, only one objective is considered, the rest being neglected. More commonly, multiple objectives are combined into a single scalar objective by linear scalarization, typically using a fixed weight vector $\boldsymbol{w}$, as seen in \cite{9701330, 9209109, 9437338, 9455139, 9905595}. However, this approach implicitly assumes that the desired trade-off between objectives, encoded by $\boldsymbol{w}$, is known in advance, an assumption that does not always hold in practice. Furthermore, when the preferred trade-off changes, the solution derived using the original $\boldsymbol{w}$ becomes suboptimal, requiring fine-tuning or complete retraining (see Fig.~\ref{fig:target_front_1}).
Alternative strategies attempt to overcome this limitation by training multiple models, each optimized for a different trade-off configuration \cite{felten2024multiobjectivereinforcementlearningbased}. Although this allows flexibility in choosing trade-offs after training, it significantly increases memory consumption and computational cost (see Fig.~\ref{fig:target_front_2}).
In our previous work \cite{uavdatacollectionmorl}, we proposed MOSAC, a multi-objective extension of the discrete Soft Actor-Critic (SAC) algorithm. MOSAC is capable of learning, in a single training run, a near-optimal policy for any desired trade-off between data collection and energy consumption (see Fig.~\ref{fig:target_front_3}). However, due to its feedforward architecture and static state representation, MOSAC also suffers from limited generalization to environments with changing parameters.

In this work, we address the trade-off between data collection and energy consumption in UAV path planning within urban environments. Building upon our previous work \cite{uavdatacollectionmorl}, we propose an algorithm that solves any trade-off vector $\boldsymbol{w}$ without requiring fine-tuning. Unlike prior approaches, this study also tackles the critical challenge of parameter generalization, enabling the learned policy to perform robustly in previously unseen environments. This contributes to improved sample efficiency, a major bottleneck in real-world applications of RL \cite{DBLP:journals/corr/abs-1904-12901}. To the best of our knowledge, this is the first approach that combines multi-objective learning with parameter generalization, leveraging an attention-based architecture inspired by inductive bias techniques for generalization \cite{Kirk_2023}.

\subsection{Contributions and Organization}
Our primary contributions are summarized as follows:

\begin{itemize}
  \item We formulate a single UAV-based wireless data harvesting task as a multi-objective path planning problem, aiming to maximize data collection while minimizing energy consumption. To address the absence of prior knowledge about the preferred trade-off between these objectives, we model the problem as a Multi-Objective Markov Decision Process (MOMDP), enabling flexible policy learning across varying trade-off preferences.
 \item To solve the formulated MOMDP, we adopt our MOSAC method, which is well-suited for environments where the dynamics of wireless channels are unknown. In particular, our approach uses low-dimensional feature vectors rather than high-dimensional spatial maps, significantly reducing model complexity and improving training efficiency.
\item To this end, we design an attention-based architecture that is inherently permutation-invariant with respect to the set of IoT devices. Unlike conventional multi-objective RL approaches that concatenate the preference vector $\boldsymbol{w}$ to the state, we introduce it as a dedicated input token to the self-attention encoder, allowing for more flexible and effective conditioning on the desired trade-off.
\item We conduct a comprehensive evaluation of our approach, comparing it against state-of-the-art DRL algorithms and a carefully designed greedy baseline. Our results demonstrate the superiority of our method in zero-shot generalization, both within training scenarios and, more importantly, in unseen environments with varying numbers of devices, spatial configurations, data volumes, and initial battery levels.
 \item Finally, we provide an in-depth analysis of the attention scores, revealing that the robustness and generalization capabilities of our algorithm arise from its ability to selectively focus on task-relevant features while effectively filtering out information that is irrelevant to the specific objective being optimized.
\end{itemize}

\noindent
The remainder of this paper is organized as follows. Section~\ref{sec:problem} introduces the system model, detailing the UAV dynamics, the communication channel model, and the formulation of the multi-objective optimization problem. In Section~\ref{sec:momdp}, we model the system as a MOMDP and discuss various state representation strategies. Section~\ref{sec:mosac} presents our attention-based multi-objective DRL algorithm. Simulation results and performance evaluations are provided in Section~\ref{sec:experiments}.

\section{System model}

\label{sec:problem}
In this section, we present the models corresponding to each component of our problem: the physical environment, the UAV, the IoT devices, and the communication framework, which closely follow our previous work \cite{uavdatacollectionmorl}. 

\subsection{Environment and UAV model}
\myitem{Environment.} We consider an urban environment discretized into a rectangular grid world of size $L \times W \in \mathbb{N}^2$, step size $c$, where $\mathcal{E}$ denotes the set of all possible positions. Such a discretization is common in RL applications, as it reduces both the state space and the number of possible actions, thus improving training efficiency and exploration. The urban environment includes buildings of varying shapes and heights, surrounded by alleys of varying widths. Let $\mathcal{H} \subset \mathcal{E}$ denote the set of positions occupied by the buildings, and $h_{x,y}$ the height of the building at location $(x,y) \in \mathcal{H}$.  We represent the grid world as a matrix $\boldsymbol{E}$, where each element $E_{x,y}$ is defined as:
\begin{equation}
    \label{eq:height_map}
    E_{x,y} = \begin{cases}
                        h_{x,y} \quad \text{if $(x,y) \in \mathcal{H}$,} \\
                        0 \quad \text{otherwise.}
                    \end{cases}
\end{equation}
$H$ denotes the height of the tallest building on the map. The environment has designated start and terminal zones in $\mathcal{E} \setminus \mathcal{H}$, denoted respectively as $\boldsymbol{p}^{s} = (x^{s}, y^{s})$ and $\boldsymbol{p}^{f} = (x^{f}, y^{f})$.

\myitem{Devices.} The $K$ static IoT devices are randomly deployed at ground level within the environment. The location of device $k$ is denoted by $\boldsymbol{p}^k = (x^k, y^k) \in \mathcal{E} \setminus \mathcal{H}$, ensuring that no device is placed on a building. At the start of the mission, each device $k$ is assigned an initial data load $D_0^k \in \mathbb{R}$ to be collected. As the mission progresses, the remaining data at time $t$ is denoted by $D_t^k \in \mathbb{R}$.

\myitem{UAV model.} In this work, we consider a single UAV navigating the described environment at a constant altitude $h \in \mathbb{R}$, such that $h > H$, where $H$ is the height of the tallest building. This setup ensures collision avoidance and reduces the complexity of the problem to a 2D trajectory optimization task. The UAV operates above the urban landscape, which reflects real-world prototyping scenarios while capturing the core challenge of the data collection problem: finding the optimal horizontal path to approach IoT devices or designated landing zones.
The mission duration is discretized into $T$ time slots of equal length $\Delta t$, indexed by $t \in {1, \ldots, T}$. The UAV's objective is to collect data from the $K$ static IoT devices located at ground level. The UAV begins the mission at the start position $\boldsymbol{p}^{s}$ with a fully charged battery $b_0 = B \in \mathbb{N}$, and the mission terminates only when it successfully reaches the terminal position $\boldsymbol{p}^{f}$. We assume that the battery level is discretized for modeling purposes. At time $t$, the UAV is characterized by its 2D position $\boldsymbol{p}_t = (x_t, y_t) \in \mathbb{R}^2$ and its remaining battery level $b_t \in \mathbb{N}$. We assume that, within a single time slot, the UAV can either hover in place or move a distance equal to $c$ (the cell size) in one of the four cardinal directions. Accordingly, the UAV's action set is defined as:
\begin{equation}
\label{eq:action_def}
\mathcal{A} =\Biggl\{\underbrace{\begin{bmatrix}
           0 \\
           0 
         \end{bmatrix}}_{\text{hover}},
\underbrace{\begin{bmatrix}
           0 \\
           c 
         \end{bmatrix}}_{\text{north}}, \underbrace{\begin{bmatrix}
           -c \\
           0 
         \end{bmatrix}}_{\text{west}}, \underbrace{\begin{bmatrix}
           0 \\
           -c 
         \end{bmatrix}}_{\text{south}}, \underbrace{\begin{bmatrix}
           c \\
           0 
         \end{bmatrix}}_{\text{east}} \Biggl\}.
\end{equation}
Thus, at time $t$, after selecting an action $\boldsymbol{a}_t \in \mathcal{A}$, the position of the UAV evolves according to:
\begin{equation}
\label{eq:uav_position_evolution}
  \boldsymbol{p}_{t+1} =  \boldsymbol{p}_{t} + \boldsymbol{a}_t.
\end{equation}
Note that certain action restrictions are enforced when the battery level is low or when the UAV is positioned at the edge of the map, as detailed in Section~\ref{subsec:mdp_conversion}. Battery consumption in UAVs generally depends on several factors, including flight mode, UAV design (e.g., payload, battery capacity, maximum speed), and environmental conditions such as wind and weather~\cite{beigi2022overviewdroneenergyconsumption}. In line with our discretized battery model and action space, we assume that moving in any of the four cardinal directions consumes one unit of battery, while hovering consumes half a unit. This assumption reflects the widely observed fact that hovering typically requires less energy than translational movement. Accordingly, the evolution of the UAV’s battery level is given by:
\begin{equation}
\label{eq:battery_evolution}
  b_{t+1} =
    \begin{cases}
       b_{t} - 0.5 & \text{if } \boldsymbol{a}_{t} = \text{hover}, \\
       b_{t} - 1 & \text{otherwise}, \\
    \end{cases}       
\end{equation}
and $e_t  = b_{t} - b_{t+1}$ denotes the energy consumed at $t$.

\subsection{Channel model and data collection}
To enable data collection from the IoT devices during the mission, a wireless connection must be established between the UAV and the devices. In each time interval $t \in {1, \ldots, T}$, the amount of data collected successfully by the UAV depends on the quality of the communication link with each device. Specifically, we write the Signal-to-Noise Ratio (SNR) between the UAV and device $k$ at time $t$ as follows~\cite{chen2023modelaidedfederatedreinforcementlearning}:
\begin{equation}
 \label{eq:SNR_definition}
  \text{SNR}^k_{t} = \frac{P10^{0.1g^k_{t}} }{\sigma^2},
\end{equation}
where $P$ denotes the transmit power, $\sigma^2$ is the power of the additive white Gaussian noise (AWGN) at the receiver, and $g^k_{t}$ represents the channel gain (in dB) between the UAV and the device $k$ at time $t$.  The channel gain is defined as $g^k_{t} = \beta_z - \alpha_z \log_{10}(d^k_t) + \eta_{t,z}$. As defined above, the channel gain depends primarily on the Euclidean distance between the UAV and device $k$, given by $d^k_t = | \boldsymbol{p}^k - \boldsymbol{p}_t |_2$. It also incorporates log-normal shadowing, modeled as $\eta_{t,z} \sim \mathcal{N}(0, \sigma_z^2)$, and sampled at every movement of the UAV, the path loss exponent $\alpha_z$, and the average channel gain $\beta_z$ at a reference distance of 1 meter. Following the channel models in \cite{9437338,7997333,9162896,9174950,9504602}, we did not account for additional randomness in the gain.
Critically, the channel gain also depends on the presence of obstacles, specifically buildings, between the UAV and the device $k$. This is captured by the link condition indicator $z \in \{\text{LoS}, \text{NLoS}\}$, which distinguishes between line-of-sight (LoS) and non-line-of-sight (NLoS) scenarios. The link condition $z$ affects the path loss exponent $\alpha_z$, the reference gain $\beta_z$, and the shadowing variance $\sigma_z^2$, thus playing a central role in the determination of communication quality.
Consequently, the maximum achievable data rate is the following:
\begin{equation}
\label{eq:rate_definition}
R^k_{t} = \log_2(1+\text{SNR}^k_{t}),
\end{equation}
where the rate is assumed to remain constant throughout the time slot (block fading model). Since the amount of data that can be collected is limited by the remaining data available at device $k$, the effective data rate is given by:
\begin{equation}
\label{eq:actual_rate_definition}
C^k_{t} = 
    \begin{cases}
      R^k_{t} & \text{if $D^k_t \geq R^k_{t}\Delta t$ , }\\
        D^k_{t}/\Delta t & \text{otherwise}.\\
    \end{cases}
\end{equation}
We assume that only one device can be scheduled for data collection during each time slot. The scheduling decision depends on the SNR: the device with the highest SNR is selected for communication. As a result, the scheduling decision is determined automatically and is not part of the UAV's action space.
We define the scheduling variable $\delta^k_t$ such that $\delta^k_t =1$ if the device $k$ is scheduled, and $\delta^k_t =0$ otherwise. It follows:
\begin{equation}
\label{eq:scheduling_variable_condition}
\sum_{k=1}^{K}\delta^k_{t} \leq 1, \quad \forall t \geq 0.
\end{equation}
It also follows that the data volume remaining at device $k$ evolves according to: $D_{t+1}^k = D_{t}^k -\delta^k_t C^k_{t} \Delta t$.

\subsection{Problem definition}
Given the environment described previously, the UAV model and dynamics, and the communication model, we now formulate the multi-objective path planning problem for data collection and energy consumption:
\begin{subequations}
\begin{alignat}{2}
&\!\max_{ \pi}        &\qquad& \left(\sum^{T}_{t=1}\sum^{K}_{k=1}\delta^k_{t}C^k_t\Delta t, \quad -\sum^{T}_{t=1}e_t\right)\label{eq:optProb}\\
&\text{s.t.} &      & \boldsymbol{p}_1 = \boldsymbol{p}^s, \boldsymbol{p}_T = \boldsymbol{p}^f ,\label{eq:constraint1}\\
&                  &      & b_T \geq 0,\label{eq:constraint2}\\
&                  &      & (\ref{eq:uav_position_evolution}), (\ref{eq:battery_evolution}), (\ref{eq:scheduling_variable_condition}).\label{eq:constraint3}
\end{alignat} \label{eq:wholeOptProb}
\end{subequations}
We aim to find policies that simultaneously optimize two conflicting objectives: maximizing the total volume of data collected and minimizing the battery consumption over the course of the mission. Note that we used the negative of the battery consumed in order to have a maximization problem for both objectives. Constraint~(\ref{eq:constraint1}) ensures that the UAV starts the mission from the designated start zone and finishes at the terminal zone. Constraint~(\ref{eq:constraint2}) guarantees that the UAV reaches the terminal zone before depleting its battery, while constraint~(\ref{eq:constraint3}) represents environmental rules and constraints, such as obstacle avoidance and movement feasibility.
The complexity of the problem arises from several factors:
\begin{enumerate}[a:]
    \item $\mathcal{NP}$-hardness of the underlying path planning problem;
    \item Multi-objective trade-off, where maximizing data collection often implies higher battery consumption;
    \item Model uncertainty, particularly due to incomplete or no knowledge of the communication environment;
    \item Stochastic initialization, including randomness in key parameters such as the number of devices $K$, their positions $p^k$, the initial data volumes $D^k_0$ for each device $k$, and the UAV's initial battery capacity.
\end{enumerate}

\noindent
To address these challenges, we propose an attention-based MORL algorithm, presented in Section~\ref{sec:mosac}.

\section{Multi-Objective Markov Decision Process}

\label{sec:momdp}
We reformulate the problem defined in \ref{sec:problem} as a MOMDP to enable its solution via RL. In addition, we introduce several state representation strategies that significantly influence both the design of the learning architecture and the performance.

\subsection{Definition}
\label{subsec:mdp_definition}
A MOMDP \cite{10.1007/11672142_26} is defined as a tuple $(\mathcal{S}, \mathcal{A}, \mathcal{P}, R, \gamma, \mu, f)$, where $\mathcal{S}$ is the state space, $\mathcal{A}$ is the action space, $\mathcal{P}$ is the transition probability function, $R: \mathcal{S} \times \mathcal{A} \xrightarrow{} \mathbb{R}^M$ is the vector-valued reward function, with $M$ denoting the number of objectives, $\gamma \in [0, 1)$ is the discount factor, $\mu$ is the initial state distribution, and $f$ is the utility function that aggregates or evaluates the trade-offs among the multiple objectives.
Given a policy $\pi$, defined as a mapping from states to actions, the multi-objective value function $\boldsymbol{V}^{\pi}$ is the expected discounted return obtained by following policy $\pi$, and is defined as:
\begin{equation}
\label{eq:moq_value_definition}
\boldsymbol{V}^{\pi} = \mathbb{E}_{\pi,s \sim \mu} \left [ \sum^{\infty}_{t=0} \gamma^t \boldsymbol{R}_{t} | s_0 = s\right].
\end{equation}
$\pi$ is said to Pareto-dominate another policy $\pi'$ (denoted $\pi \succ_P \pi'$) if  $\pi$ yields equal or higher value than $\pi'$ in all objectives and strictly higher value in at least one objective. Formally, this is expressed as $(\forall m: V_m^{\pi} \geq V_m^{\pi'} ) \wedge (\exists  m: V_m^{\pi} > V_m^{\pi'})$.
Accordingly, the objective is to identify the set of non-dominated policies, denoted by $\Pi^*$, defined as:
\begin{equation}
\label{eq:pareto_dominance_value}
\Pi^* = \{\pi^* | \nexists \pi': \boldsymbol{V}^{\pi'} \succ_P \boldsymbol{V}^{\pi^*} \}.
\end{equation}
A preference vector $\boldsymbol{w}$ is introduced to express the relative importance of each objective, where $\boldsymbol{w} \in \mathbb{R}^M$ satisfies $\sum_m w_m = 1$ and $w_m \geq 0, \forall m$. This vector is typically used to compare multi-objective value functions via linear scalarization, defined as:
$f(\boldsymbol{V}^{\pi},\boldsymbol{w}) = \boldsymbol{w}^T\boldsymbol{V}^{\pi}$. 
For a given preference vector $\boldsymbol{w}$, if no other policy yields a higher scalarized utility than $\pi$, then the corresponding value $\boldsymbol{V}^{\pi}$ is considered non-dominated with respect to that preference~\cite{Roijers_2013}.

\subsection{MOMDP conversion}
\label{subsec:mdp_conversion}
\myitem{State space.} The system state is defined by the environment map, the UAV state, and the states of the IoT devices. The environment map is static and represented by the matrix $\boldsymbol{E}$, as defined in \eqref{eq:height_map}. At time $t$, the UAV state is given by $s^{\text{UAV}}_t = (b_t^{\text{SC}}, b_t, d_{t,x}, d_{t,y})$, where $b_t$ is the current battery level, $b_t^{\text{SC}}$ is the minimum battery required to reach the destination (terminal zone), and $d_{t,x} = x^f - x_t$, $d_{t,y} = y^f - y_t$ are the relative distances to the terminal position in the $x$ and $y$ directions.
The state of each IoT device $k$ is represented as $s^k_t = (d^k_{t,x}, d^k_{t,y}, D^k_t, \text{SNR}^k_t, \rho^k_t)$, where $d^k_{t,x} = x^k - x_t$ and $d^k_{t,y} = y^k - y_t$ are the relative distances between the UAV and the device $k$, $D^k_t$ is the remaining data to collect, $\text{SNR}^k_t$ is the corresponding SNR, and $\rho^k_t$ is a binary indicator equal to $1$ if the device is reachable (i.e., if $\text{SNR}^k_t \geq \Sigma$), and $0$ otherwise.
 
\myitem{Action Space.} The action space corresponds directly to the set of UAV actions defined in \eqref{eq:action_def}. At each time step, the UAV selects one action from this set, such as hovering or moving to an adjacent cell, which determines its motion within the environment.
Following the execution of an action, the MOMDP transitions to a new state according to the environment dynamics described in Section~\ref{sec:problem}.

\myitem{Vector reward.} 
The reward provides an instantaneous measure of the long-term objectives defined in \eqref{eq:optProb} and is given by:
\begin{equation}
\label{eq:reward_definition}
\boldsymbol{r}_t = \left[\sum_{k=1}^{K} \delta^k_t C^k_t \Delta t, -e_t \right],
\end{equation}
where the first component represents the amount of data collected during time step $t$, and the second component captures the negative of the energy consumed.

\myitem{Safety Controller.} To reflect realistic operational constraints and enforce safety measures, a safety controller is integrated to restrict certain actions in the following scenarios:
\begin{itemize}
\item \textit{Map boundaries:} when the UAV is located at the edge of the map, any action that would result in moving outside the defined grid is masked and thus prevented.
\item \textit{Critical battery level:} when the minimum battery required to reach the destination becomes zero ($b_t^{\text{SC}} = 0$), only actions that lead directly to the terminal zone are allowed.
\end{itemize}

\subsection{State representation}
State representation plays a critical role in this problem, as it directly affects the design of the DNN architecture, the generalization capability of the learned policy, and the effective handling of the multi-objective nature of the task.

\myitem{Feature Vector (FTV).} In our previous work \cite{uavdatacollectionmorl}, we used the most conventional state representation for this type of problem:
\begin{equation}
\label{eq:state_definition_ftv}
s^{\text{FTV}}_t = [s_t^{\text{UAV}}, s_t^{1}, \ldots,  s_t^{K}].
\end{equation}
This feature vector representation, however, suffers from several important limitations: First, it imposes the use of a feedforward neural network architecture with a fixed input size. Specifically, the state vector has a dimensionality of $4 + 5K$, which prevents the agent from handling scenarios with a variable number of devices $K$. Second, the representation is sensitive to the ordering of the devices in the state vector. Permuting two devices into the defined state in \eqref{eq:state_definition_ftv} can lead the trained agent to misinterpret the state as different and produce a suboptimal or incorrect action. 
Third, as shown in~\cite{9437338}, this representation tends to perform poorly in environments with variable parameters, such as dynamic or randomly initialized device positions. It lacks the flexibility and inductive bias required to generalize effectively across such variations.
Despite these drawbacks, the feature vector framework does offer a practical advantage: it simplifies the implementation of standard deep MORL methods that rely on concatenating state and preference vectors as input.

\begin{figure}[H]
    \centering
    \includegraphics[width=0.98\linewidth]{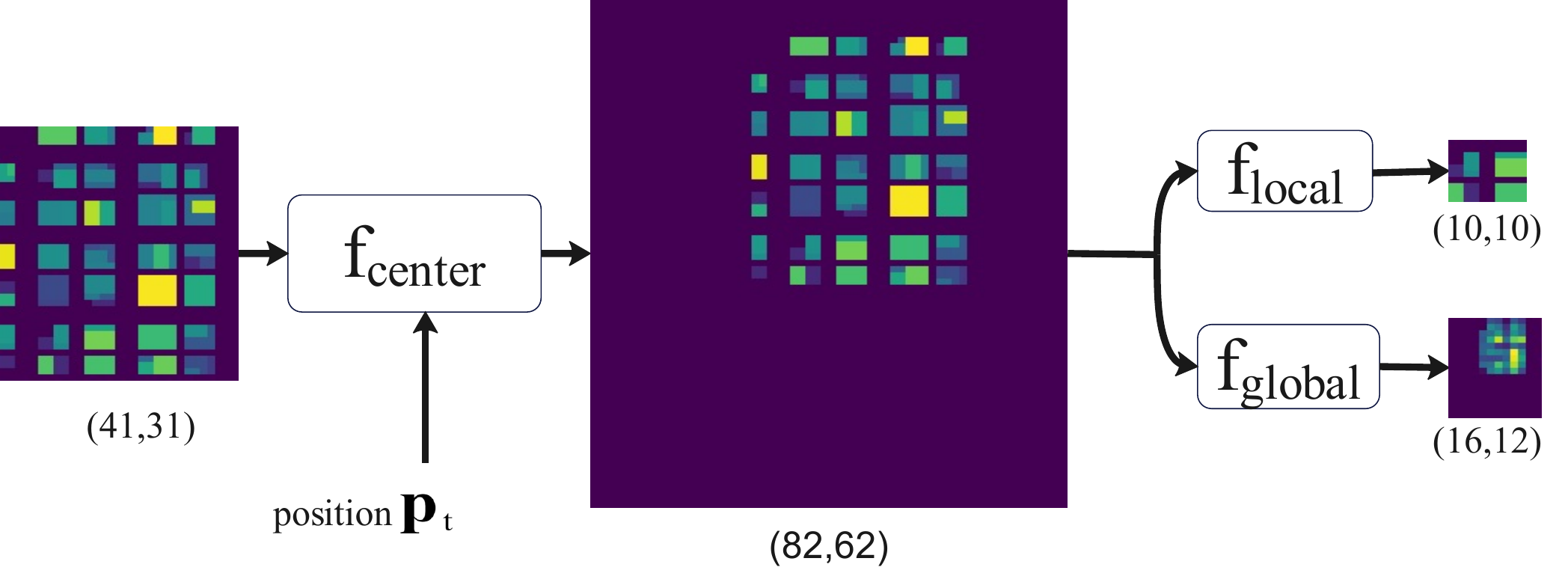}
    \caption{Map processing on matrix $\boldsymbol{E}$ with $l =10$ and $g = 5$. }
    \label{fig:cgl_map}
\end{figure}

\myitem{Global/local maps.} To address the limitations of the feature vector representation, an efficient spatial map-based representation was proposed in~\cite{Theile_2021}, which incorporates three key map-processing operations. 
The state space is defined as:
\begin{equation}
\label{eq:state_definition_cgl} 
\mathcal{S} = \underbrace{\mathbb{R}^{L\times W \times 2}}_{\text{Environment Map}} \times \underbrace{ \mathbb{R}^{L\times W}}_{\text{Devices map}}   \times \underbrace{\mathbb{B}^{L\times W}}_{\text{UAV position}}   \times \underbrace{\mathbb{N}}_{\text{Flying time}} .
\end{equation}
We adapt the spatial map-based representation to our rectangular grid environment and problem setting. The state $\boldsymbol{S}_t$ is represented as the concatenation of following $L \times W$ matrices:
\begin{itemize}
\item $\boldsymbol{E}$ containing building heights (see \eqref{eq:height_map}),
\item $\boldsymbol{Z}$: a location map with a value of $1$ at the start zone, $-1$ at the terminal zone, and $0$ elsewhere,
\item $\boldsymbol{D}_t$: a dynamic map indicating the remaining data $D^k_t$ at each IoT device position $\boldsymbol{p}^k$; all other entries are zero,
\item $\boldsymbol{P}_t$: a binary map marking the UAV's current position $\boldsymbol{p}_t$ with a $1$, and $0$ elsewhere.
\end{itemize}
\noindent 
Building on this representation, three processing functions are applied: $f_{\text{center}}$, $f^l_{\text{local}}$, and $f^g_{\text{global}}$. The centering process $f_{\text{center}}$ returns a matrix centered around the UAV position. The local mapping $f^l_{\text{local}}$ extracts a central crop of size $l \times l$, focusing on the UAV's immediate surroundings.
The global mapping $f^g_{\text{global}}$ applies average pooling over non-overlapping windows of size $g \times g$, capturing broader contextual information.
Two spatial maps are then derived using the previously defined functions:
$\boldsymbol{S}_t^{\text{CL}} = f^l_{\text{local}}(f_{\text{center}}(\boldsymbol{S}_t))$ and $\boldsymbol{S}_t^{\text{CG}} = f^g_{\text{global}}(f_{\text{center}}(\boldsymbol{S}_t))$. Refer to Fig. \ref{fig:cgl_map} for a visual illustration and to \cite{Theile_2021} for further details. This architecture imposes the use of two parallel convolutional networks (CNNs) \cite{Szegedy2014GoingDW}, whose outputs are flattened and concatenated in subsequent layers to form a latent representation. This latent vector is then augmented with the remaining flight time $b_t$. While this approach proved to be efficient in terms of performance and parameter generalization, we identify several limitations inherent to spatial map representations. In particular, they require large convolutional models and result in memory-inefficient state representations (note the sparsity of the matrices in \eqref{eq:state_definition_cgl}). Furthermore, access to the full city map $\boldsymbol{E}$ may not always be feasible in practice. Another limitation is related to multi-objective learning: within this framework, the preference $\boldsymbol{w}$ can only be incorporated after the flattening and concatenation of the CNN outputs, potentially restricting its influence on earlier spatial feature extraction stages.

\myitem{Set representation.} To address the aforementioned drawbacks, we represent the state of the problem as a set and adopt a DNN architecture that is invariant to permutations of its elements. The state is initially defined as $\{s_t^{\text{}}, s_t^{1}, \ldots,  s_t^{K} \}$.
In this context, additional components such as the preference vector $\boldsymbol{w}$ and the cropped local map $\boldsymbol{S}_t^{\text{CL}}$, corresponding to the FOV of the UAV, can also be included as elements of the set. This yields the final state representation:
\begin{equation}
\label{eq:state_definition_attention}
\boldsymbol{s}_t = \{s^{\text{UAV}}_t, s_t^{1}, \ldots,  s_t^{K}, \boldsymbol{w}, \boldsymbol{S}_t^{\text{CL}} \}.
\end{equation}
Since the attention mechanism \cite{vaswani2023attentionneed}, when used without positional encoding, is inherently permutation-invariant, we leverage it in our MORL algorithm introduced in the following section. Unlike the spatial map-based approach, the combination of set representations with attention still faces limitations related to fixed input sizes. To address this, we define a maximum number of devices $K^{max}$, analogous to the maximum sequence length used in language models \cite{vaswani2023attentionneed}. This upper bound ensures compatibility with the model architecture, and in practice, it suffices to choose $K^{max}$ large enough to cover the expected number of devices in the environment.

\begin{figure*}[!t]
    \centering
    \includegraphics[width=0.95\textwidth]{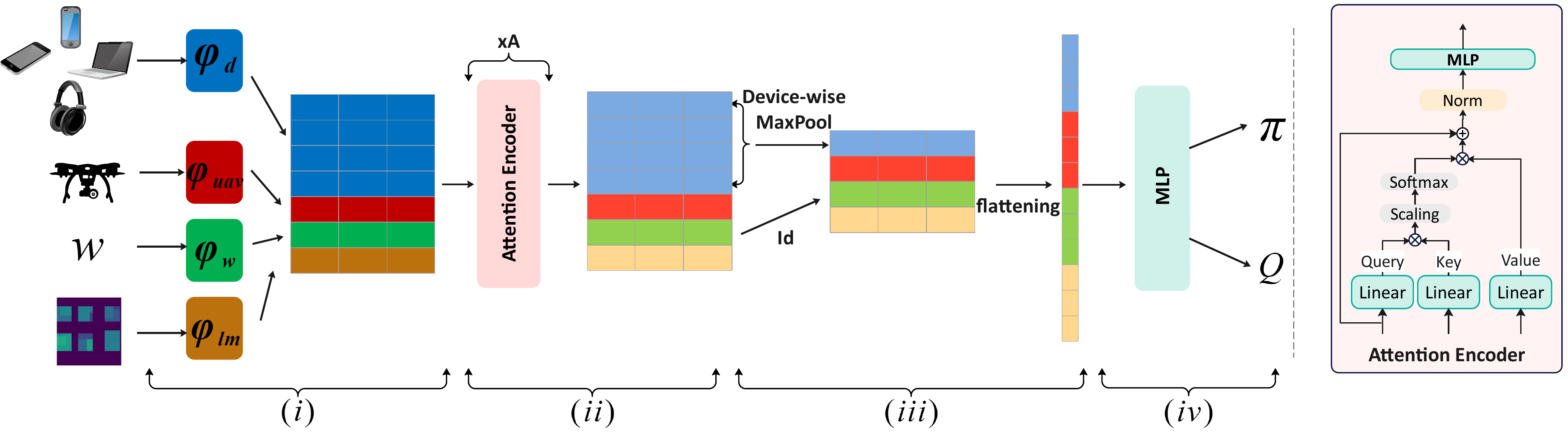}
  \hfill
  \caption{Attention-based DNN architecture used as RL agent. The projection layers $\phi$ are linear maps without positional encoding.}
  \label{fig:architecture}
\end{figure*}

\section{Multi-Objective Soft Actor-Critic (MOSAC)}

\label{sec:mosac}
\subsection{Learning algorithm}
We build on our previous work, the MOSAC algorithm \cite{uavdatacollectionmorl}, by extending the network architecture and replacing the feedforward DNN with an attention-based model that supports the set-style representation of the state. Our MOSAC algorithm has already demonstrated significant performance improvements over baseline deep MORL methods, primarily due to the exploration enhancements we introduced.
MOSAC is a multi-objective extension of the discrete Soft Actor-Critic (SAC) algorithm \cite{christodoulou2019softactorcriticdiscreteaction}, which is a model-free method that optimizes a trade-off between expected return and policy entropy. It maintains three neural networks, parameterized respectively by $\phi$, $\theta$, and $\bar{\theta}$: a policy network (estimator) $\pi_{\phi}:\mathcal{S} \rightarrow [0,1]^{|\mathcal{A}|}$, an action-value (Q-value) estimator $Q_{\theta}:\mathcal{S} \rightarrow \mathbb{R}^{|\mathcal{A}|}$, and a target action-value (Q-network) $Q_{\bar{\theta}}$ used to stabilize learning. To mitigate overestimation bias in Q-value learning, $Q_{\theta}$ is duplicated, following standard practice in actor-critic methods.

\myitem{Multi-objective Q estimator.} We extend the Q-function and the corresponding target expression to accommodate the MORL setting. The modified Q objective is defined as:
\begin{equation}
\label{eqn:j_q_multi_objective}
J_Q(\theta) = \underset{\substack{(s, a, \boldsymbol{r})\sim \mathcal{B} \\ \boldsymbol{w} \sim \mathcal{D}_{w}}}{\mathbb{E}} \left[ \frac{1}{2} \| \boldsymbol{Q}_\theta(s, a,\boldsymbol{w}) - \boldsymbol{y}(s, a, \boldsymbol{w}) \|_2^2 \right],
\end{equation}
where $\mathcal{B}$ denotes the replay buffer and $\mathcal{D}_{w}$ is a sampling distribution over the preference space. The target value $\boldsymbol{y}$ incorporates both reward and entropy, governed by a trade-off parameter $\nu$, and is defined as:
\begin{multline}
\label{eqn:y_target_multi_objective}
 \boldsymbol{y}(s, a, \boldsymbol{w})= \boldsymbol{r} + \gamma \mathbb{E}_{s'} \left[  \left( \boldsymbol{Q}_{\bar{\theta}}(s', \cdot, \boldsymbol{w}) - \right.  \right. \\  \left. \left. \nu \boldsymbol{1}_{M} \log \boldsymbol{\pi}(s', \boldsymbol{w})^T \right) \boldsymbol{\pi}(s', \boldsymbol{w}) \right],
\end{multline}
where $\boldsymbol{1}_{x}$ is the vector of size $x$ with all elements equal to $1$.

\myitem{Multi-Objective Policy.} The policy is updated by minimizing the following objective:
\begin{multline}
\label{eqn:j_policy_multi_objective}
J_\pi(\phi) = \underset{\substack{s\sim \mathcal{B} \\ \boldsymbol{w} \sim \mathcal{D}_{w}}}{\mathbb{E}}\left[ \boldsymbol{\pi}_\phi(s,  \boldsymbol{w})^T \left[ \nu \log \boldsymbol{\pi}_\phi(s,  \boldsymbol{w}) -   \right.  \right. \\  \left. \left. \boldsymbol{Q}_\theta(s, \cdot, \boldsymbol{w})^T\boldsymbol{w} \right] \right].
\end{multline}
Note that distinct action-values and policies are learned for different preference vectors $\boldsymbol{w}$, as the scalarized utility is explicitly incorporated in the objective function. Furthermore, by training on sampled preferences drawn from the distribution $\mathcal{D}_{w}$, our DNN estimators gain the ability to generalize across the entire preference space, as demonstrated in \cite{pmlr-v97-abels19a}.

\myitem{Entropy trade-off.} $\nu$ is learned to regulate policy entropy, encouraging it to match a target entropy level $H_0$.

\begin{equation}
\label{eqn:j_alpha_multi_objective}
J(\nu) = \underset{\substack{s\sim \mathcal{B} \\ \boldsymbol{w} \sim \mathcal{D}_{w}}}{\mathbb{E}}\left[ \boldsymbol{\pi}_\phi(s, \boldsymbol{w})^T \left[ -\nu \left( \log \boldsymbol{\pi}_\phi( s, \boldsymbol{w}) +  H_0\boldsymbol{1}_{|\mathcal{A}|} \right) \right]\right].
\end{equation}

\myitem{Heated-up softmax.} We significantly improved the performance of the discrete SAC algorithm by introducing tailored exploration strategies. While discrete SAC is known for its stability and sample efficiency, it can still suffer from suboptimal convergence in certain cases. This issue often arises from the challenge of balancing the trade-off between the expected discounted return and policy entropy.
To address this, we implemented a decaying entropy schedule, where the target entropy is gradually reduced from an initial value $H_0$ to a final (lower) value $H_{\text{final}}$, thereby progressively emphasizing return maximization over exploration as training advances.
In parallel, we introduced a temperature-based policy sharpening mechanism. Specifically, we added a temperature parameter $\tau$, which is linearly annealed from $\tau_0$ to $\tau_{\text{final}}$ over the course of training. This modifies the stochastic policy as follows:
\begin{equation}
\label{eqn:tau_heated_up}
\pi^{\tau}(a|s) = \frac{\exp{(l_a  / \tau)} }{\sum_i \exp{(l_i/ \tau)}},
\end{equation}
where $l_i$ denotes the learned logit associated with action $i$. This annealing mechanism initially encourages broader exploration and gradually shifts toward more deterministic behavior. 
As shown in our previous work, these dynamic exploration techniques outperform approaches with a fixed entropy target, leading to faster and more reliable convergence.

\begin{figure*}[!t]
  \centering
  \begin{subfigure}{.25\textwidth}
    \centering
    \includegraphics[width=\textwidth]{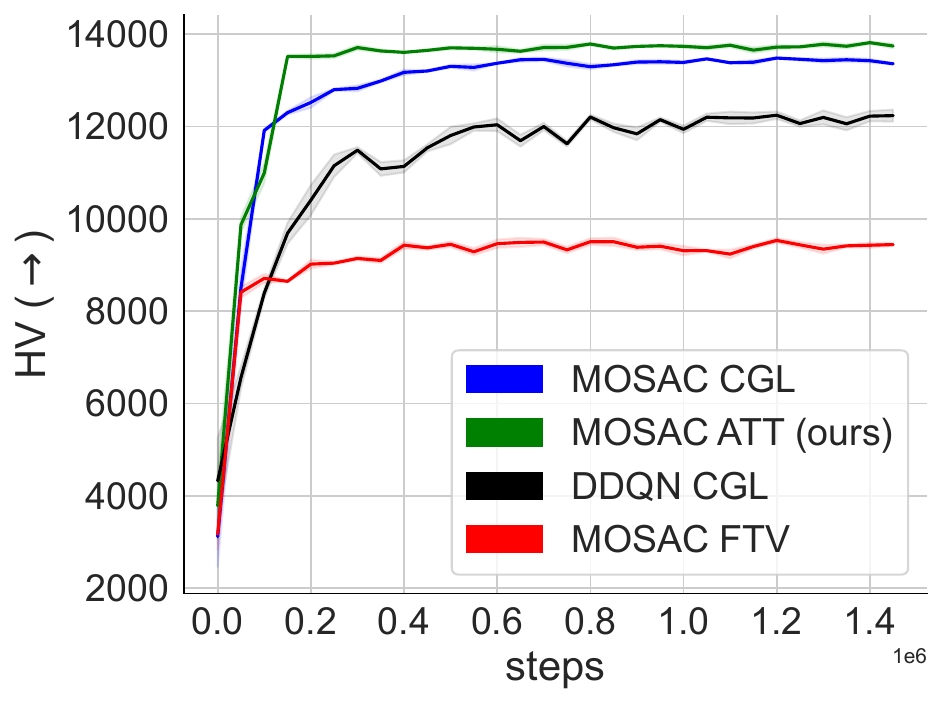}
    \caption{Hypervolume - RD}
    \label{fig:hypervolume_1}
  \end{subfigure}%
  \hfill
  \begin{subfigure}{.25\textwidth}
    \centering
    \includegraphics[width=\textwidth]{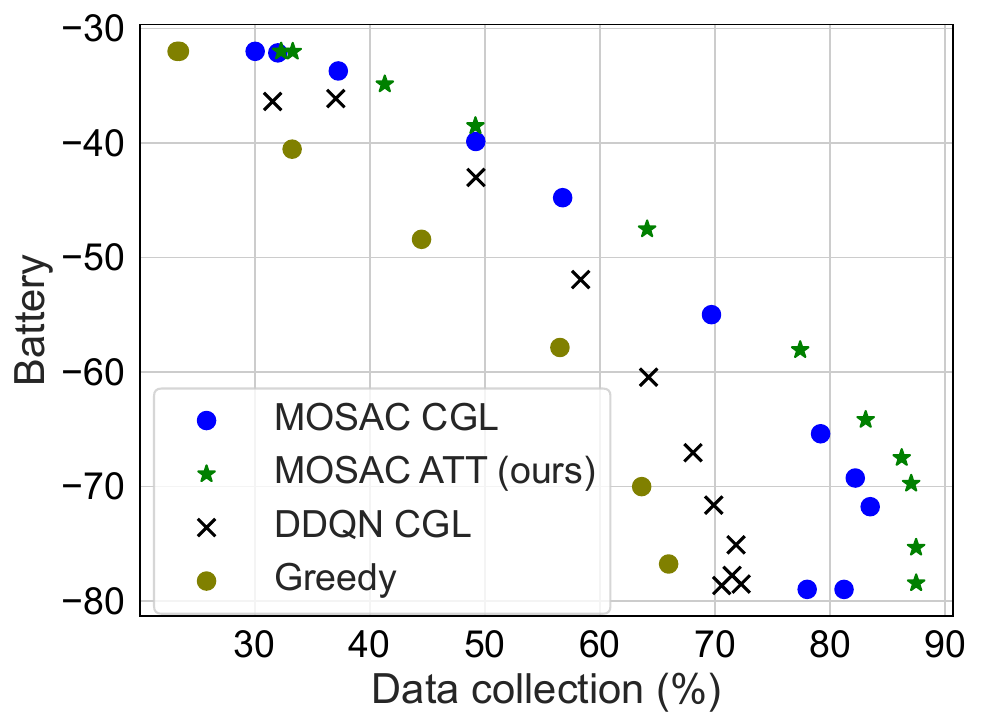}
    \caption{Pareto Front - RD}
    \label{fig:front_1}
  \end{subfigure}%
  \hfill
  \begin{subfigure}{.25\textwidth}
    \centering
    \includegraphics[width=\textwidth]{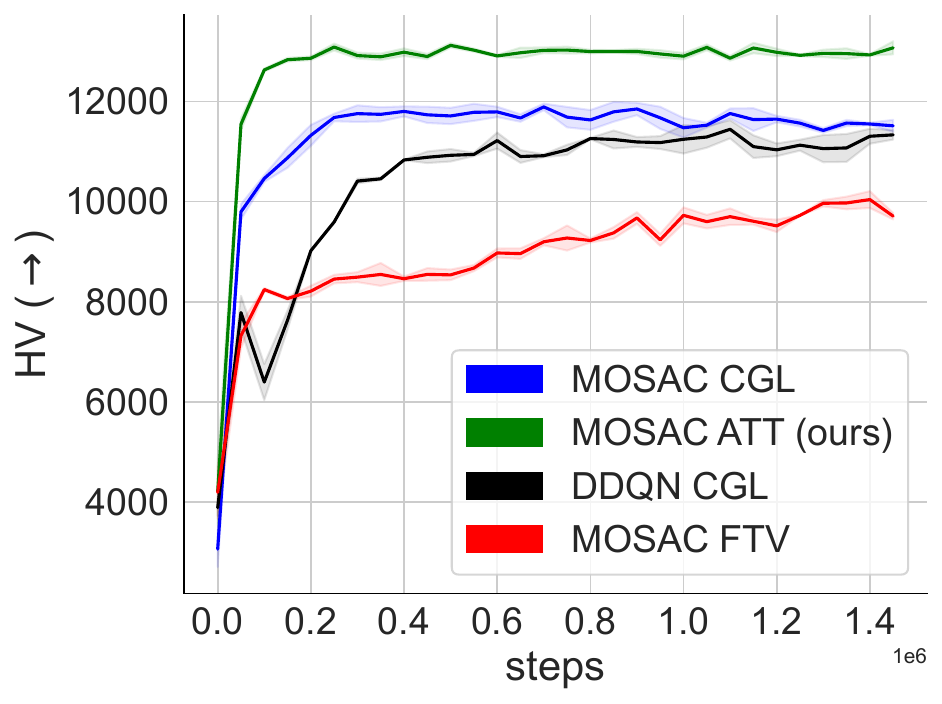}
    \caption{Hypervolume - RB}
    \label{fig:hypervolume_2}
  \end{subfigure}%
  \hfill
  \begin{subfigure}{.25\textwidth}
    \centering
    \includegraphics[width=\textwidth]{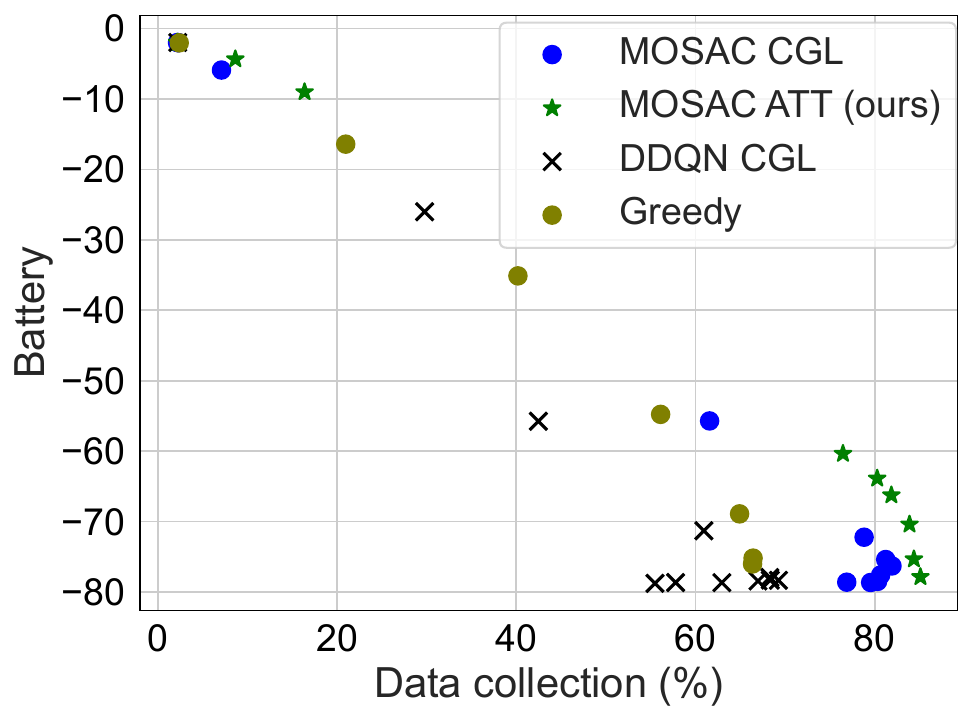}
    \caption{Pareto Front - RB}
    \label{fig:front_2}
  \end{subfigure}%
  \hfill
  \caption{Multi-objective performance averaged across $3$ seeds over training steps and Pareto fronts of final saved models.
  }
  \label{fig:trj_varios_pref}
\end{figure*} 

\subsection{Attention-based Architecture}
We propose a novel self-attention-based architecture to replace the original feedforward model. 

\myitem{Self-Attention}, popularized by \cite{vaswani2023attentionneed} in the context of Natural Language Processing (NLP), enables the model to learn relevant projections of input vectors (tokens) to generate output vectors that capture the most salient features for a given task. In our problem, the input sequence is given by ${s^{\text{UAV}}_t, s_t^{1}, \ldots, s_t^{K}, \boldsymbol{w}, \boldsymbol{S}_t^{\text{CL}}}$, which lacks inherent positional information, an ideal setting for attention mechanisms that are inherently token permutation-invariant.
Furthermore, within a multi-objective learning framework, we incorporate the preference $\boldsymbol{w}$ directly into the input sequence to guide the agent's perception of the state in accordance with the desired trade-off. For instance, when the goal is solely to minimize battery consumption, as indicated by $\boldsymbol{w} = (0, 1)$, the model should place minimal emphasis on features related to the devices in the environment map. Instead, the agent should focus primarily on its remaining flight time and the position of the terminal zone. Conversely, when data collection is prioritized, the agent should attend more strongly to the devices, including their positions and the amount of remaining data.

\myitem{Our proposed architecture} comprises four components, as shown in Fig. \ref{fig:architecture}: ($i$) Since our problem is represented as a heterogeneous set of elements, including the UAV state, device information, local map, and preference, we first apply linear projectors to transform each type of input into vectors of a common dimension. These projectors are implemented as feedforward networks and are denoted by $\phi_{\text{uav}}$, $\phi_d$, $\phi_{\text{lm}}$, and $\phi_{\boldsymbol{w}}$ for the UAV state, device information, local map, and preference, respectively. The resulting vectors serve as the equivalent of word embeddings in NLP applications. ($ii$) These embeddings, each of dimension $d$ (a tunable hyperparameter), form a matrix $X \in \mathbb{R}^{K^{max}+3,d}$ that then passes through a sequence of $A$ attention encoder layers. Each attention encoder projects its input into queries $Q$, keys $Z$, and values $V$ following:
\begin{equation*}
\label{eqn:qkv_definition}
Q = XW_Q, \quad Z = XW_Z, \text{and }  V = XW_V,
\end{equation*}
where $W_Q$, $W_Z$ and $W_V$ are trainable linear projectors in  $\mathbb{R}^{d\times d_q}$, 
$\mathbb{R}^{d\times d_q}$ and $\mathbb{R}^{d\times d_v}$  respectively.
The attention operation is written as:
\begin{equation*}
\label{eqn:attention_definition}
\text{Attention}(Q,Z,V) =\text{softmax}(\frac{QZ^T}{\sqrt{d_q}})V,
\end{equation*} 
with $d_q$ the dimension of the queries and keys, $\sqrt{d_q}$ a scaling term and the softmax operation ensures that attention weights are positive and sum to one. The complete attention encoder is depicted on the right side of Fig. 3 and generates contextualized representations that capture dependencies among the different elements in the input sequence. ($iii$) Before proceeding to the prediction step, the contextualized vectors must be aggregated. To ensure device-wise permutation invariance, we apply a MaxPool operation over the device-related vectors. A key advantage of this pooling step is the dimensionality reduction it provides: the device-specific portion of the sequence (in blue in the figure) is compressed from a size of $K^{\text{max}} \times d$ to $1 \times d$. As a result, the size of the concatenated latent vector becomes independent of the number of devices in the environment. ($iv$) Finally, the aggregated latent vector is fed into a Multi-Layer Perceptron (MLP), which outputs the multi-objective policy $\pi$ and the corresponding action-value vector $\boldsymbol{Q}$.

\section{Simulations}
\label{sec:experiments}

\subsection{Simulation setup}
\setcounter{table}{0}

\begin{figure*}[!t]
  \centering
  \begin{subfigure}{.28\textwidth}
    \centering
    \includegraphics[width=\textwidth]{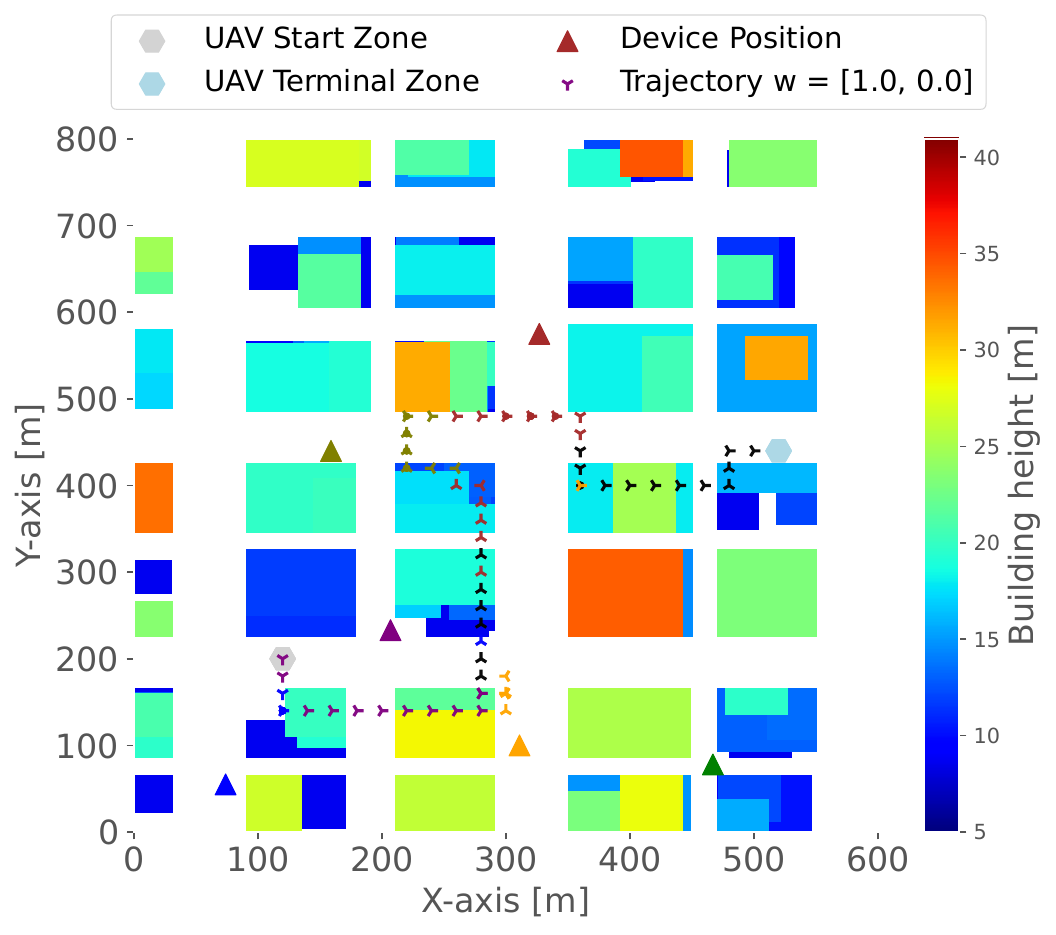}
    \vspace*{-6mm}
    \caption{$\boldsymbol{w} = (1,0)$, $\boldsymbol{J}_{\boldsymbol{w}}^{\pi} = (77\%,-78.5)$}
    \label{fig:traj_collection_1}
  \end{subfigure}%
  \hfill
  \begin{subfigure}{.28\textwidth}
    \centering
    \includegraphics[width=\textwidth]{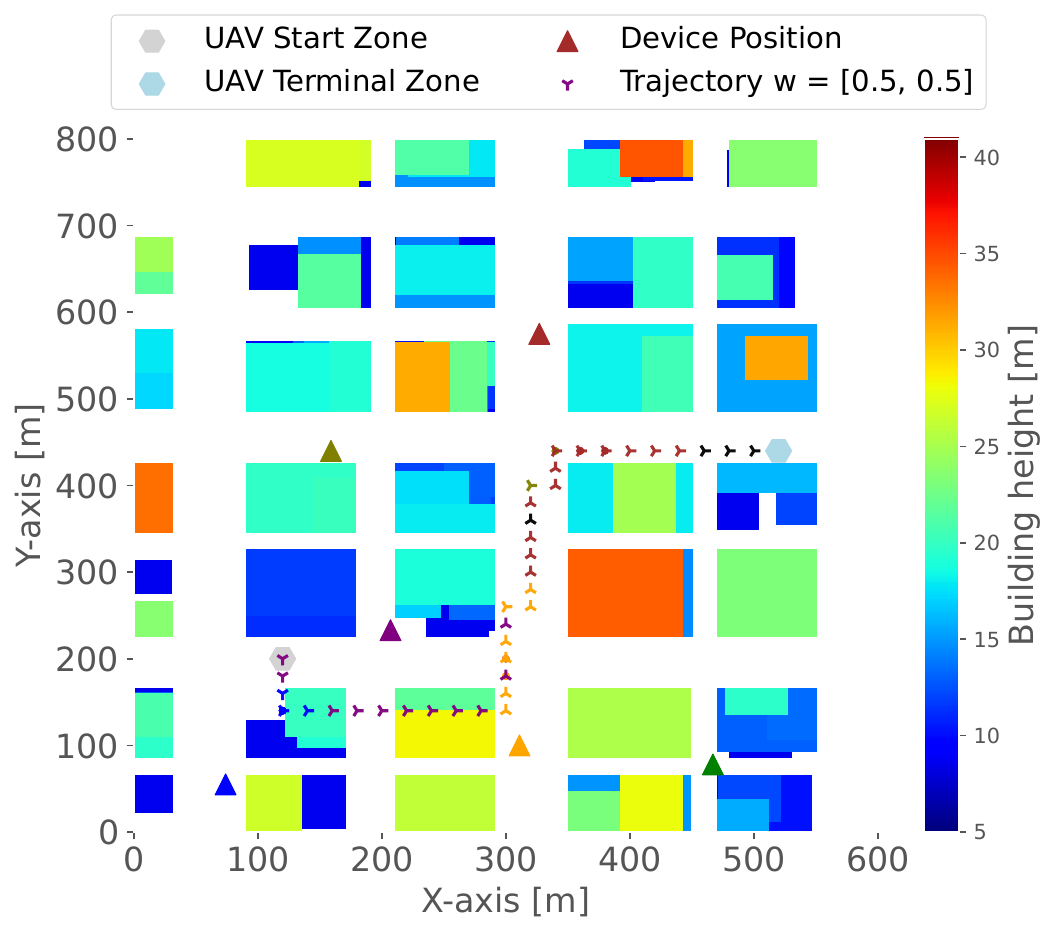}
    \vspace*{-6mm}
    \caption{$\boldsymbol{w} = (0.5,0.5)$, $\boldsymbol{J}_{\boldsymbol{w}}^{\pi} = (68\%,-48)$}
    \label{fig:traj_middle_1}
  \end{subfigure}%
  \hfill
  \begin{subfigure}{.28\textwidth}
    \centering
    \includegraphics[width=\textwidth]{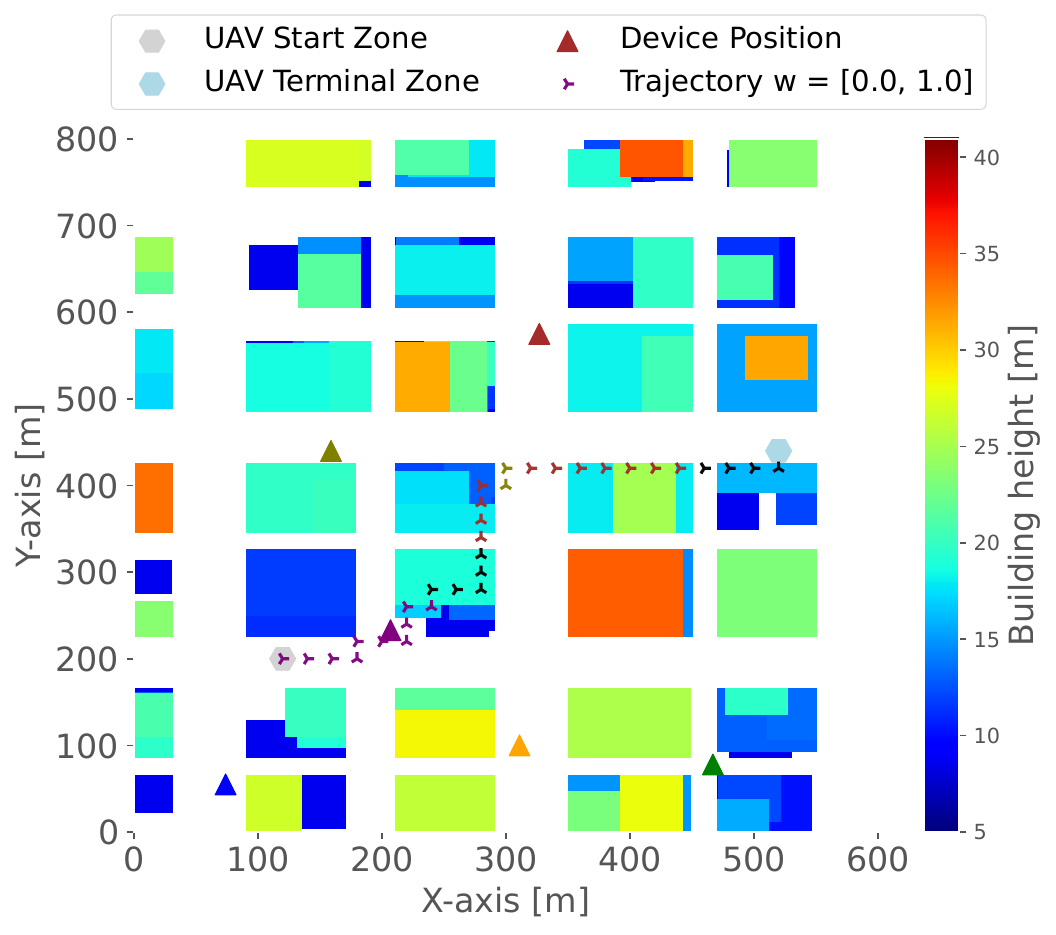}
    \vspace*{-6mm}
    \caption{$\boldsymbol{w} = (0,1)$, $\boldsymbol{J}_{\boldsymbol{w}}^{\pi} = (25\%,-32)$}
    \label{fig:traj_battery_1}
  \end{subfigure}%
  \hfill
  \begin{subfigure}{.28\textwidth}
    \centering
    \includegraphics[width=\textwidth,trim={0 0 0 1.9cm},clip]{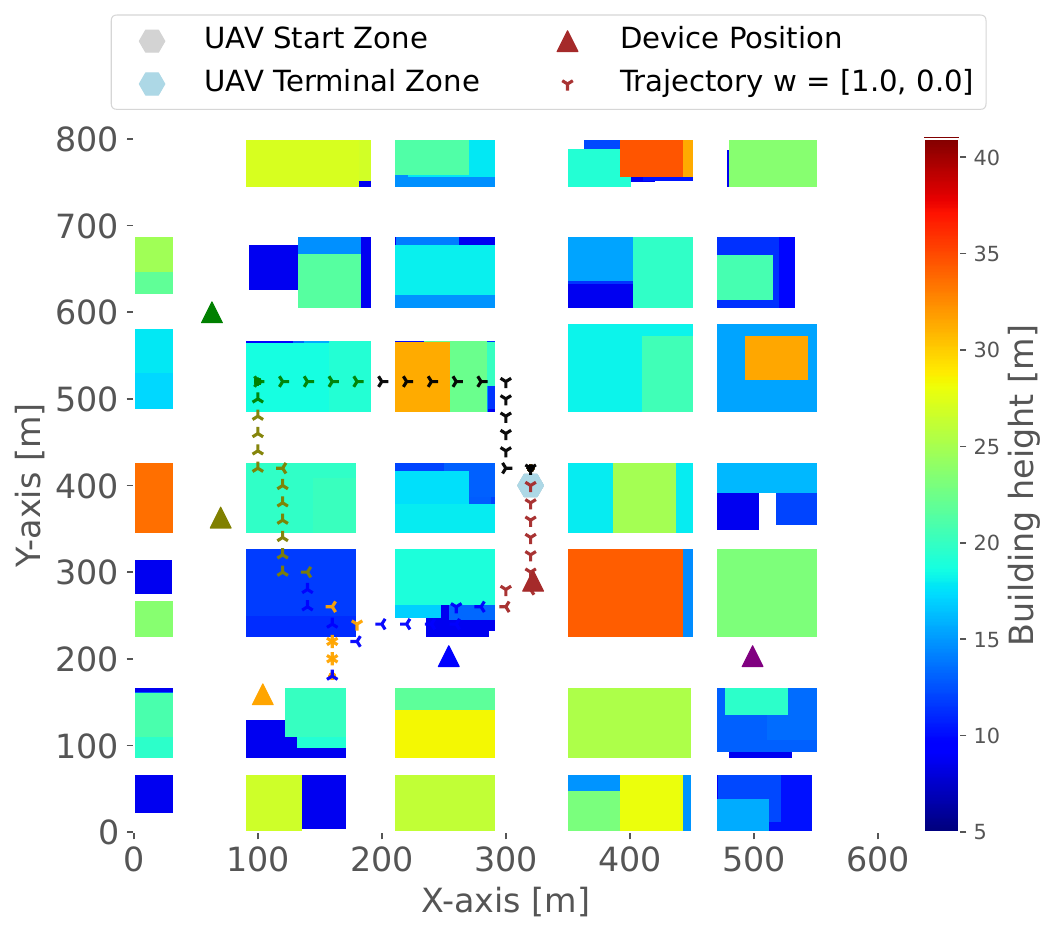}
    \vspace*{-6mm}
    \caption{$\boldsymbol{w} = (1,0)$, $\boldsymbol{J}_{\boldsymbol{w}}^{\pi} = (84\%,-78.5)$}
    \label{fig:traj_collection_2}
  \end{subfigure}%
  \hfill
  \begin{subfigure}{.28\textwidth}
    \centering
    \includegraphics[width=\textwidth,trim={0 0 0 1.9cm},clip]{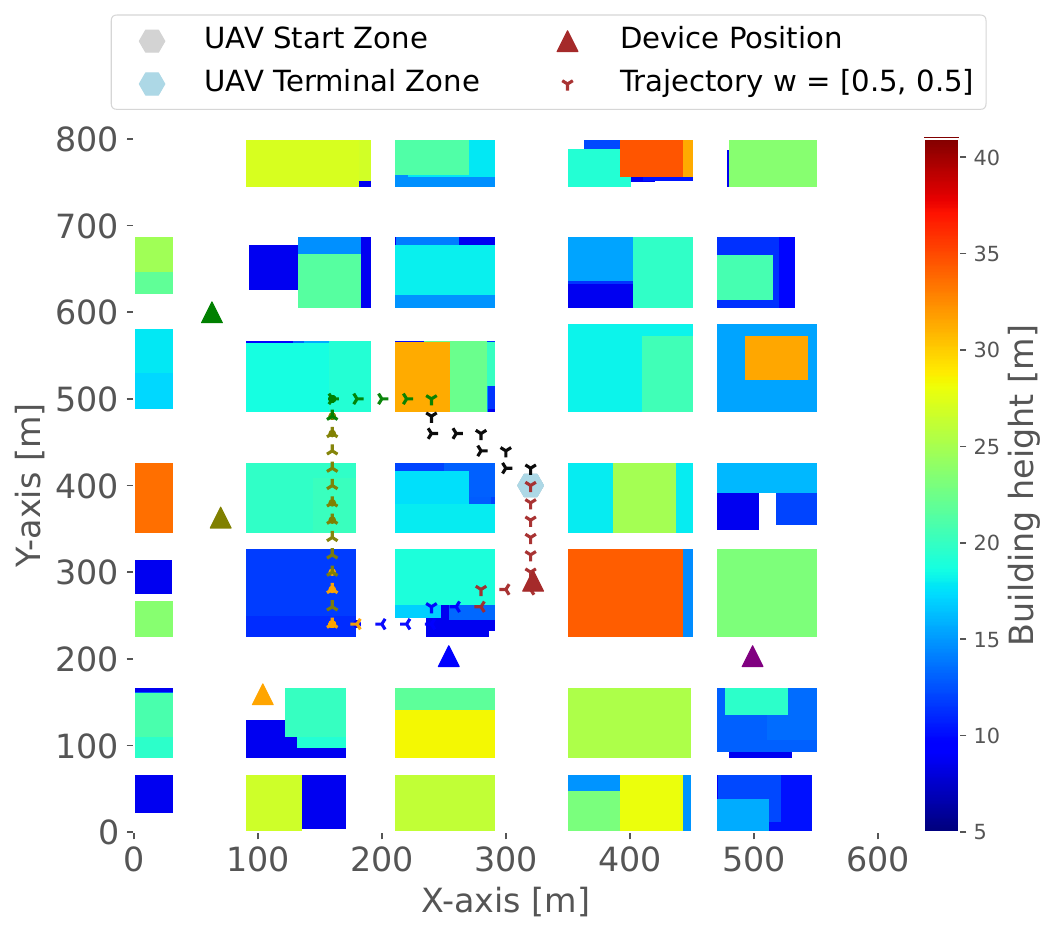}
    \vspace*{-6mm}
    \caption{$\boldsymbol{w} = (0.5,0.5)$, $\boldsymbol{J}_{\boldsymbol{w}}^{\pi} = (79\%,-59)$}
    \label{fig:traj_middle_2}
  \end{subfigure}%
  \hfill
  \begin{subfigure}{.28\textwidth}
    \centering
    \includegraphics[width=\textwidth,trim={0 0 0 1.9cm},clip]{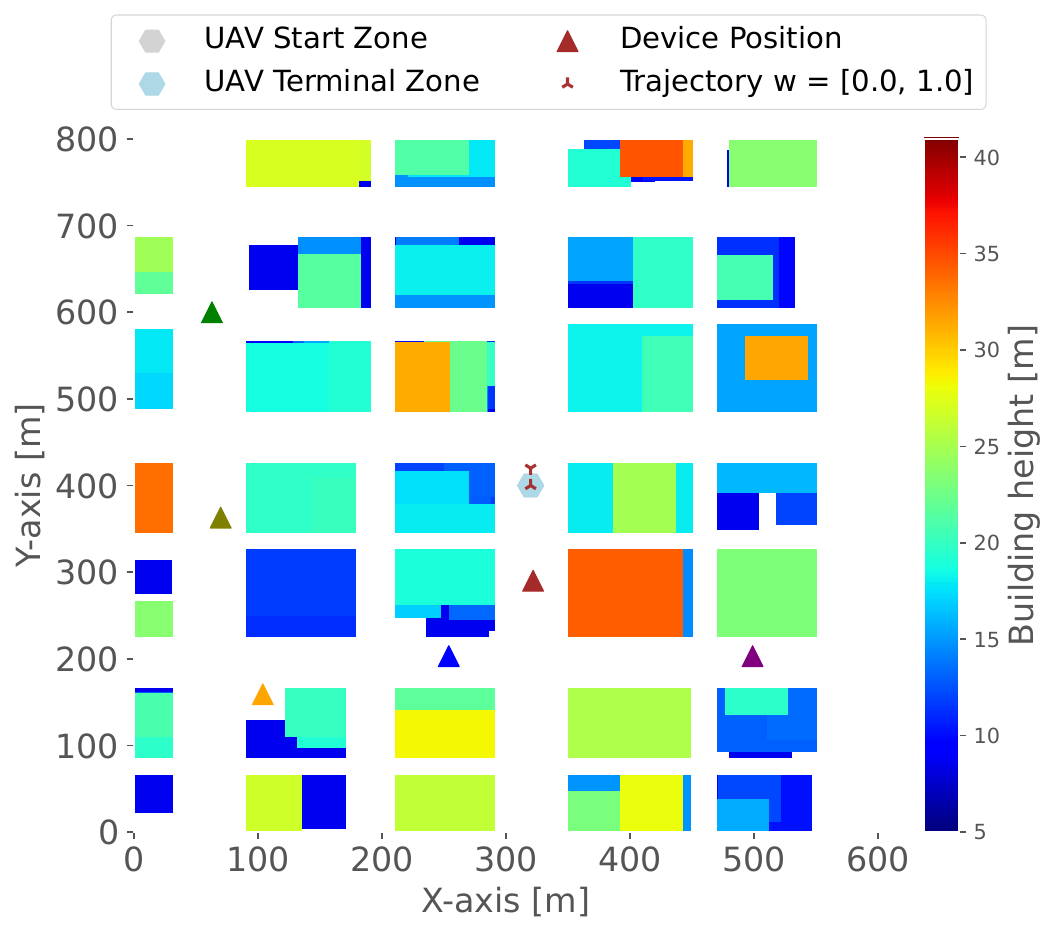}
    \vspace*{-6mm}
    \caption{$\boldsymbol{w} = (0,1)$, $\boldsymbol{J}_{\boldsymbol{w}}^{\pi} = (2\%,-2)$}
    \label{fig:traj_battery_2}
  \end{subfigure}%
  \hfill
  \caption{Examples of MOSAC-ATT trajectories for three different preferences in the RD (a, b, c) and RB (d, e, f) scenarios.
  }
  \label{fig:trj_varios_pref_1}
\end{figure*}

Our study aims to develop a multi-objective algorithm capable of generating different policies according to user-defined trade-offs between competing objectives, while demonstrating generalization across varying settings of the environment. Specifically, we adopt a training–evaluation framework where the algorithm is trained in a single, fixed environment and subsequently evaluated across multiple unseen settings.
To facilitate this, we design two simulation scenarios distinguished by the spatial configuration of the start and terminal zones. In the ``Reach Destination" (RD) scenario, the start and terminal zones are fixed and located at distinct positions (see Fig. \ref{fig:traj_collection_1}). In contrast, the ``Return Base" (RB) scenario features coinciding start and terminal zones (see Fig. \ref{fig:traj_collection_2}).

In both scenarios, the first objective is to explore the environment map and collect data from scattered devices, while the second objective is to minimize battery usage by finding the shortest path to the terminal zone. The environment maps are grids of size $(800m \times600m)$, and are discretized with a step size $c=20$m. The UAV operates at an altitude of $h=60$m, with the tallest buildings reaching a height of $H=50$m. We use the same signal propagation parameters as in \cite{uavdatacollectionmorl}: the transmit power is $P = 36$ dBm, the SNR threshold is $\Sigma = 5$dB, and the propagation parameters are $\alpha_{\text{LoS}}=2.5$, $\alpha_{\text{NLoS}}=3.04$, $\beta_{\text{LoS}}=-30$, $\beta_{\text{NLoS}}=-35$, 
$\sigma^2_{\text{LoS}}=2$, $\sigma^2_{\text{NLoS}}=5$. To evaluate algorithm performance in both scenarios, we conduct experiments on separate training and test sets. Training is carried out with $K=6$ devices and a fixed UAV battery capacity of $B=80$ units. While the number of devices and battery level remain constant during training, variability is introduced by randomizing the device positions and initial data volumes. At the beginning of each training episode, for all $k \in \{1,\ldots,6\}$, device positions $p^k$ are randomly sampled such that each pair is separated by at least $100$m, and initial data amounts $D^k_0$ are sampled uniformly from the interval $[5000,8000]$. 
For testing, we intentionally introduce greater variability to assess the algorithms' robustness to previously unseen scenarios. Specifically, we vary the number of devices (between $3$ and $10$) and the UAV's initial battery level. Each algorithm is trained using three different random seeds for a total of $1.5$ million steps, with evaluations conducted every $20,000$ steps. To ensure fair comparisons across all training runs, we generate and fix $100$ environment realizations (with $K=6$) for evaluation purposes.
After training, the final saved models are used to evaluate generalization to unseen conditions. For each value of $K \in \{3,5,6,9,10\}$, we again generate and fix $100$ distinct environment realizations to maintain consistency and fairness across algorithm evaluations.

\subsection{Evaluation Baselines and Metrics}
\noindent
The following benchmarks are considered alongside our proposed approach:

\myitem{MOSAC-ATT} utilizes the architecture described in Section~\ref{sec:mosac}. Input features are embedded into vectors of dimension $d = 64$, processed through $A=2$ attention encoder layers, each employing $4$ attention heads. All linear layers within the model consist of $128$ neurons. The maximum number of devices is fixed at $K^{max} = 12$. If the number of devices exceeds the set limit, the $K^{max}$ devices with highest SNR can be naively selected in the state, as commonly done in scheduling problems. In addition, given a fixed number of devices, $K^{max}$ does not affect performance since the empty device tokens are masked in the attention encoder.

\myitem{DDQN-CGL} is the CNN-based Double Deep Q-Network (DDQN) \cite{vanhasselt2015deepreinforcementlearningdouble} introduced in \cite{Theile_2021}. It leverages centered local and global (CGL) map representations, using a local crop size of $l=10$ and a global scaling factor of $g=5$. We extend the original architecture by concatenating a preference vector to the latent representation obtained after flattening the convolutional output. The input maps are processed through two convolutional layers, each with $16$ channels and $(5\times5)$ kernels. Following the concatenation step, a two-layer MLPs with $256$ neurons per layer is used to produce both the policy and value outputs.

\myitem{MOSAC-CGL} integrates the CGL paradigm into the MOSAC framework, utilizing the same convolutional structure and preference concatenation scheme as in DDQN-CGL. This design allows us to isolate the performance gains attributable solely to our proposed architecture.

\myitem{MOSAC-FTV} corresponds to the algorithm presented in our previous work. It employs the same feedforward architecture and hyperparameter settings as originally proposed.

\myitem{GREEDY} follows a heuristic strategy based on shortest-path decisions. The UAV navigates to the nearest device, hovers to collect its data, and then proceeds either to the next closest device or directly to the terminal zone, depending on the chosen parameterization.

\begin{figure*}[t]
  \centering
  \begin{subfigure}{.28\textwidth}
    \centering
    \includegraphics[width=\textwidth]{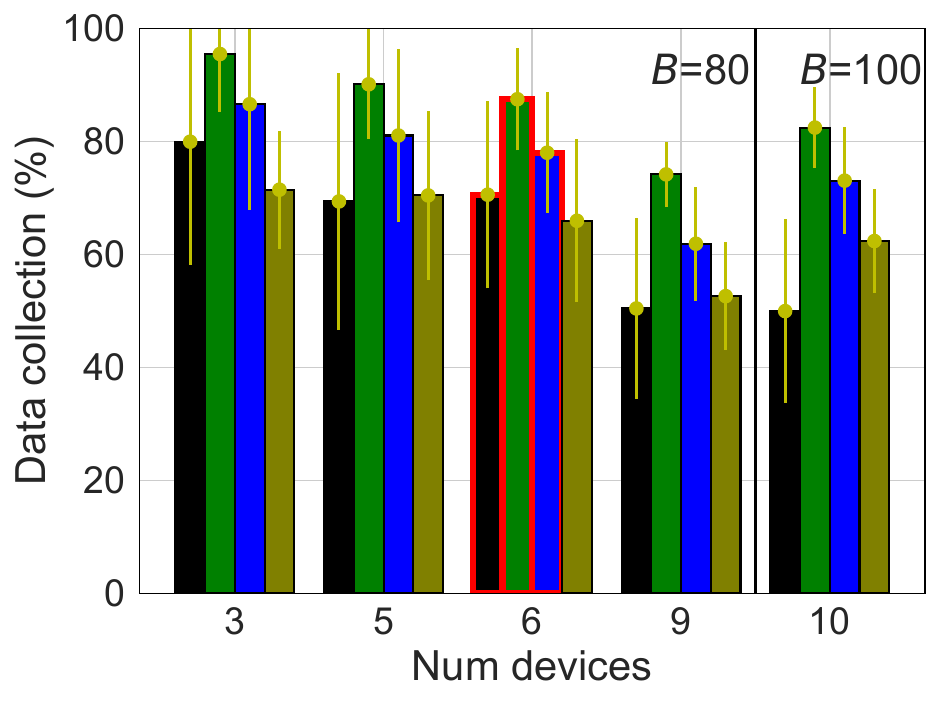}
    \caption{RD - Data collection $\boldsymbol{w} = (1,0)$}
    \label{fig:gen_collection_1}
  \end{subfigure}%
  \hfill
  \begin{subfigure}{.28\textwidth}
    \centering
    \includegraphics[width=\textwidth]{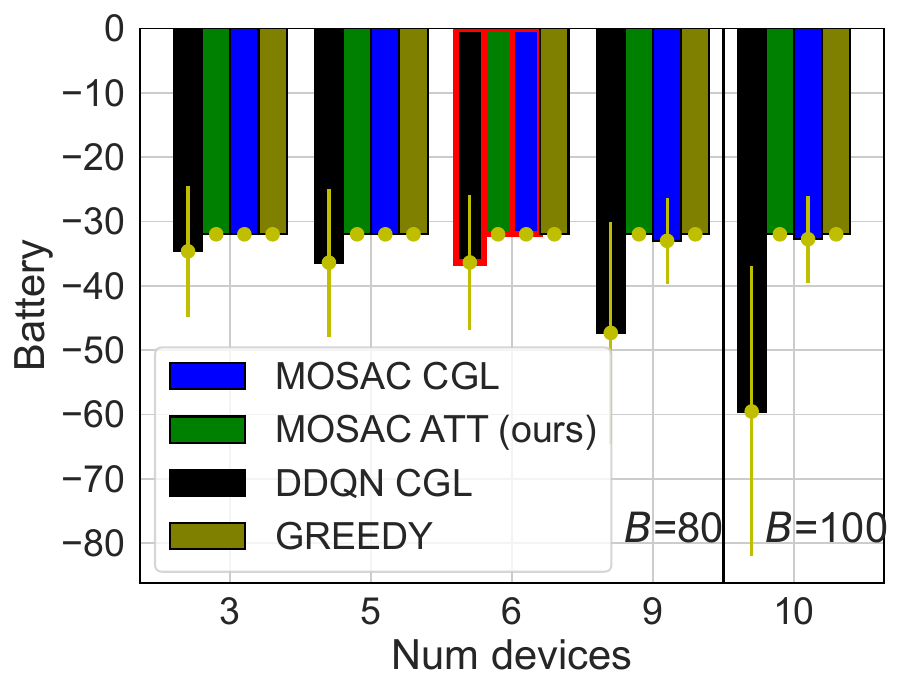}
    \caption{RD - Battery  $\boldsymbol{w} = (0,1)$}
    \label{fig:gen_battery_1}
  \end{subfigure}%
  \hfill
  \begin{subfigure}{.28\textwidth}
    \centering
    \includegraphics[width=\textwidth]{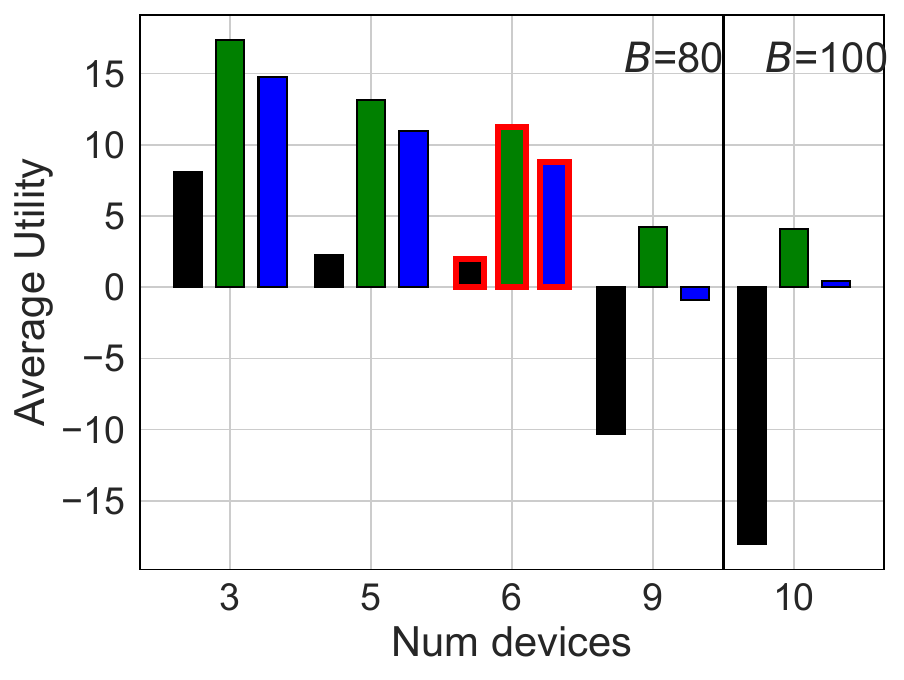}
    \caption{RD - Averaged scalarized utility}
    \label{fig:gen_utility_1}
  \end{subfigure}%
  \hfill
  \begin{subfigure}{.28\textwidth}
    \centering
    \includegraphics[width=\textwidth]{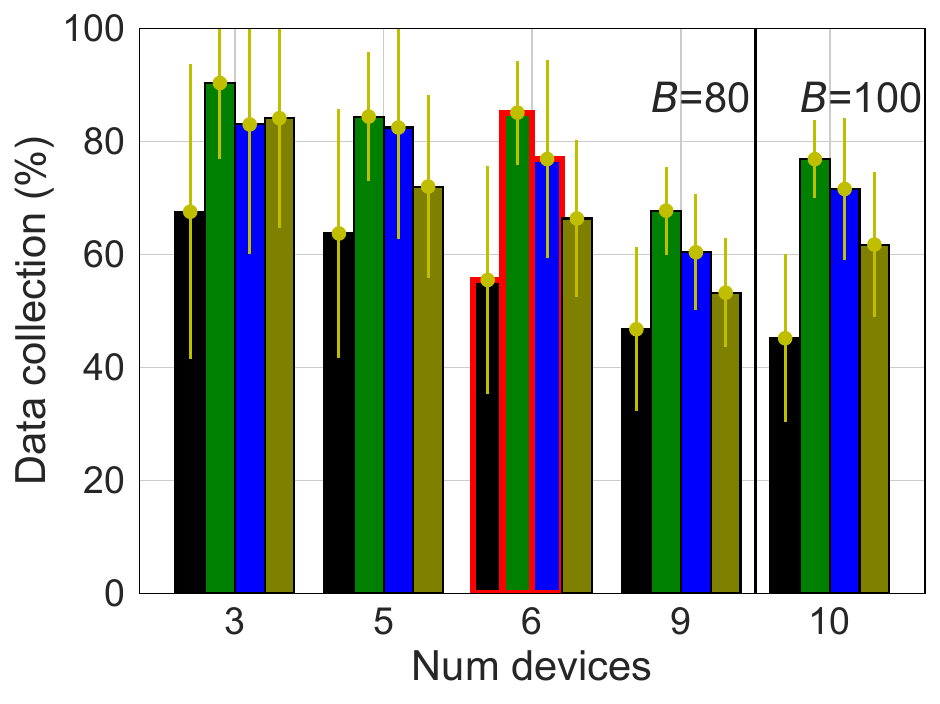}
    \caption{RB - Data collection  $\boldsymbol{w} = (1,0)$}
    \label{fig:gen_collection_2}
  \end{subfigure}%
  \hfill
  \begin{subfigure}{.28\textwidth}
    \centering
    \includegraphics[width=\textwidth]{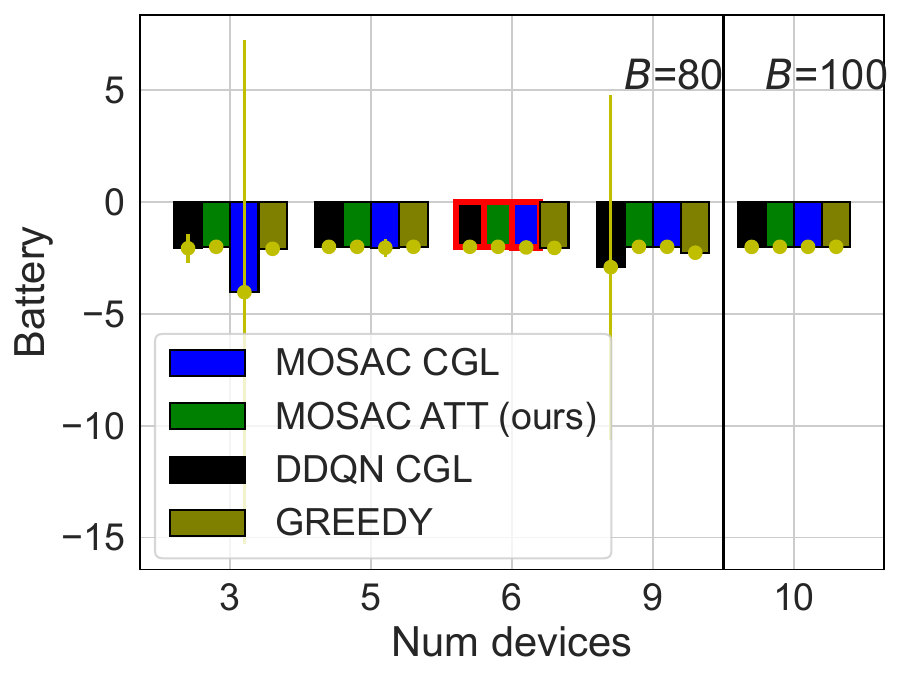}
    \caption{RB - Battery $\boldsymbol{w} = (0,1)$}
    \label{fig:gen_battery_2}
  \end{subfigure}%
  \hfill
  \begin{subfigure}{.28\textwidth}
    \centering
    \includegraphics[width=\textwidth]{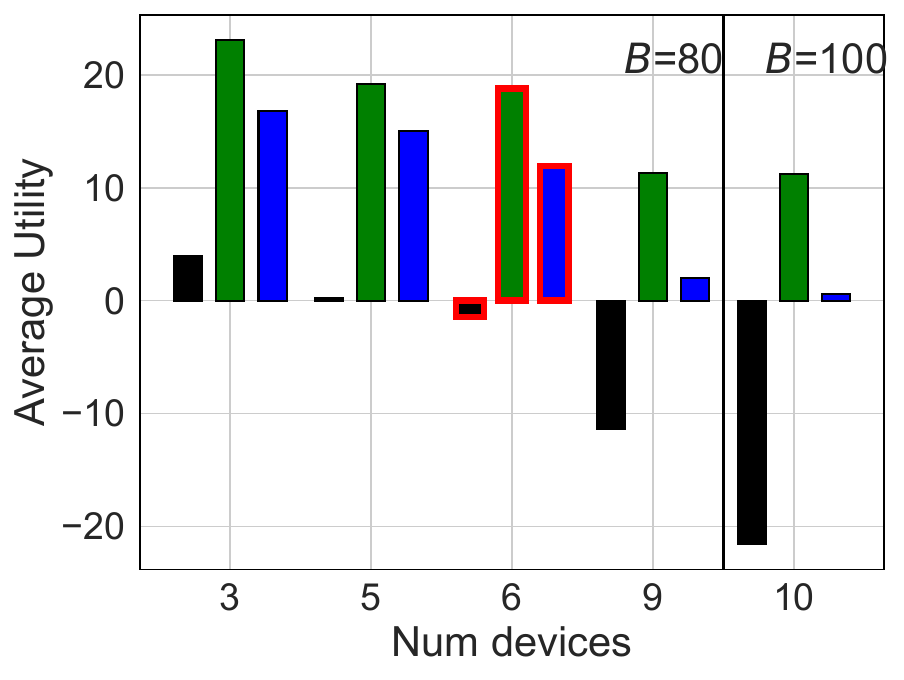}
    \caption{RB - Averaged scalarized utility}
    \label{fig:gen_utility_2}
  \end{subfigure}%
  \hfill
  \caption{Generalization to unseen settings in RD and RB scenarios (training setting highlighted in red).
  }
  \label{fig:generalization_1}
\end{figure*}

\noindent
All MORL algorithms are trained using a learning rate of $l_r = 3 \times 10^{-4}$, a batch size of $32$, a replay buffer size of $100,000$, a discount factor $\gamma = 0.99$, and $J = 3$ preference vectors sampled per update step. For MOSAC, in particular, we set the entropy coefficient schedule parameters as $H_0 = 0.6H_{\text{max}}$, $H_{\text{final}} = 0.3H_{\text{max}}$, with the temperature parameters $\tau_0 = 5$ and $\tau_{\text{final}} = 1.5$, decayed linearly over $100,000$ steps. The temperature annealing parameters were carefully selected based on our previous work \cite{uavdatacollectionmorl}, in order to outperform fixed target-entropy settings.
After training, MORL policies can be conditioned on arbitrary preferences between objectives. For evaluation, we define a test preference set $\mathcal{W}_{\text{test}} = \left\{ \left( \frac{i}{10}, 1 - \frac{i}{10} \right) \,\middle|\, i = 0, 1, \ldots, 10 \right\}$, which uniformly spans the entire preference space. This results in a set of $11$ distinct policies, denoted as $J^{\pi}$, per test environment. Given $100$ environment realizations per test condition, this setup leads to a total of $1,100$ evaluations per algorithm.
Performance is assessed using two criteria: (i) the average of the objective metrics defined in eq.~(\ref{eq:optProb}) across all test environments; (ii) the hypervolume (HV) metric~\cite{pratical_guide_morl}, which quantifies the volume of the space dominated by the set $J^{\pi}$ relative to a reference point, with higher values indicating better overall performance.
The hypervolume is defined as $HV(\boldsymbol{J}^{\text{ref}},J^{\pi} )=\bigcup\limits_{\boldsymbol{J} \in J^{\pi}} \text{Volume}(\boldsymbol{J}^{ref},\boldsymbol{J})$, where the reference point is set to $\boldsymbol{J}^{\text{ref}} = (-1, -200)$.

\subsection{Training performances, Pareto Front and computation}
The evaluation steps serve a dual purpose: they track the evolution of training performance and also provide a limited measure of generalization, given that the $100$ fixed environment realizations used for evaluation are unlikely to have been encountered during training. Figures~\ref{fig:hypervolume_1} and~\ref{fig:hypervolume_2} show the average hypervolume over training steps for the RD and RB scenarios, respectively.
The results indicate that MOSAC-ATT consistently outperforms all other algorithms in terms of multi-objective performance, training stability, and sample efficiency in both scenarios. Notably, across all random seeds, MOSAC-ATT reliably converges to a higher hypervolume than its counterparts after approximately $150,000$ steps.
Interestingly, in the RD scenario, most of the performance gain over DDQN-CGL appears to stem from the learning algorithm itself rather than the architectural improvements, as MOSAC-CGL achieves performance levels close to those of MOSAC-ATT. In contrast, in the RB scenario, the performance gap between MOSAC-ATT and MOSAC-CGL is more pronounced, highlighting the additional benefit provided by the attention-based architecture in that setting.
Figures~\ref{fig:front_1} and~\ref{fig:front_2} display the Pareto fronts averaged over the $100$ environment realizations for the RD and RB scenarios, respectively. These plots demonstrate that the GREEDY baseline is generally dominated by the RL algorithms across the preference space.
The results also show that for the extreme preference $\boldsymbol{w}=(0,1)$, all RL algorithms consistently learn shortest-path policies to the terminal zone, resulting in battery usage of approximately $32$ units in the RD scenario and $2$ units in the RB scenario, regardless of the specific environment realization. Conversely, when $\boldsymbol{w}=(1,0)$, the MOSAC-ATT algorithm achieves the highest data collection performance, with average coverage rates of approximately 87\% in the RD scenario and 85\% in the RB scenario.
It is also worth noting that the Pareto fronts obtained by the RL algorithms in the RB scenario are relatively sparse. This sparsity is likely a consequence of the limitations of linear scalarization, which is known to struggle in identifying solutions in concave or linear regions of the true Pareto front~\cite{Das1997ACL}.
Figure~\ref{fig:trj_varios_pref_1} illustrates the behavior of the policy learned by MOSAC-ATT across different preference vectors, shown for a single randomly selected environment realization in each of the two scenarios. The trajectories reveal that as the preference for data collection increases, the UAV actively moves closer to the IoT devices. This adaptive behavior demonstrates the model’s ability to adjust to varying objective trade-offs without requiring additional training.
\noindent
Finally, a critical point addressed by the CGL method in the context of spatial map representation is the reduction of deep neural network (DNN) model size. This consideration is especially important for embedded systems, which demand compact and efficient models due to limited computational resources. Our proposed architecture further advances this goal by reducing the number of trainable parameters by approximately 55\%, from $165,829$ in MOSAC-CGL to just $91,781$, while maintaining competitive performance. On the other hand, we note that, on an Intel Xeon Gold 6278C CPU (4 cores), the average inference time of MOSAC-CGL is $1.3$ms against $2.0$ms for  MOSAC-ATT, i.e. the latency slightly increased. We consider this computational overhead to be acceptable in comparison to the notable gains of our algorithm in generalization and robustness to unseen cases. In addition, this inference time range is substantially lower than the time required to travel a cell $c=20$m at velocity $20$m/s.

To measure our model's robustness, we evaluate its obtained data collection performance in an environment with additional fading. We augment the gain $g^k_{t}$ with a fading term $20\log_{10}(|h_t|)$, where $|h_t|$ is Rayleigh-distributed with scale $1$, and collect the performance averaged over the $100$ test environments repeated $30$ times. The results are recorded in the following table:
\begin{table}[H]
\renewcommand{\arraystretch}{1.5}
\caption{Robustness to randomness.}
\centering
\begin{tabular}{|l|l|l|}
\hline
\textbf{ } & \textbf{Regular} & \textbf{With fading} \\
\hline
Mean (\%) & 87.49 & 87.23 \\
Std & 0.16 & 0.28 \\
\hline
\end{tabular}
\end{table}
As the mean and standard deviation are barely impacted, we conclude that the learned policy is robust to the added fading.

\subsection{Generalization to unseen settings}

\begin{figure*}[!t]
  \centering
  \begin{subfigure}{.32\textwidth}
    \centering
    \includegraphics[width=\textwidth]{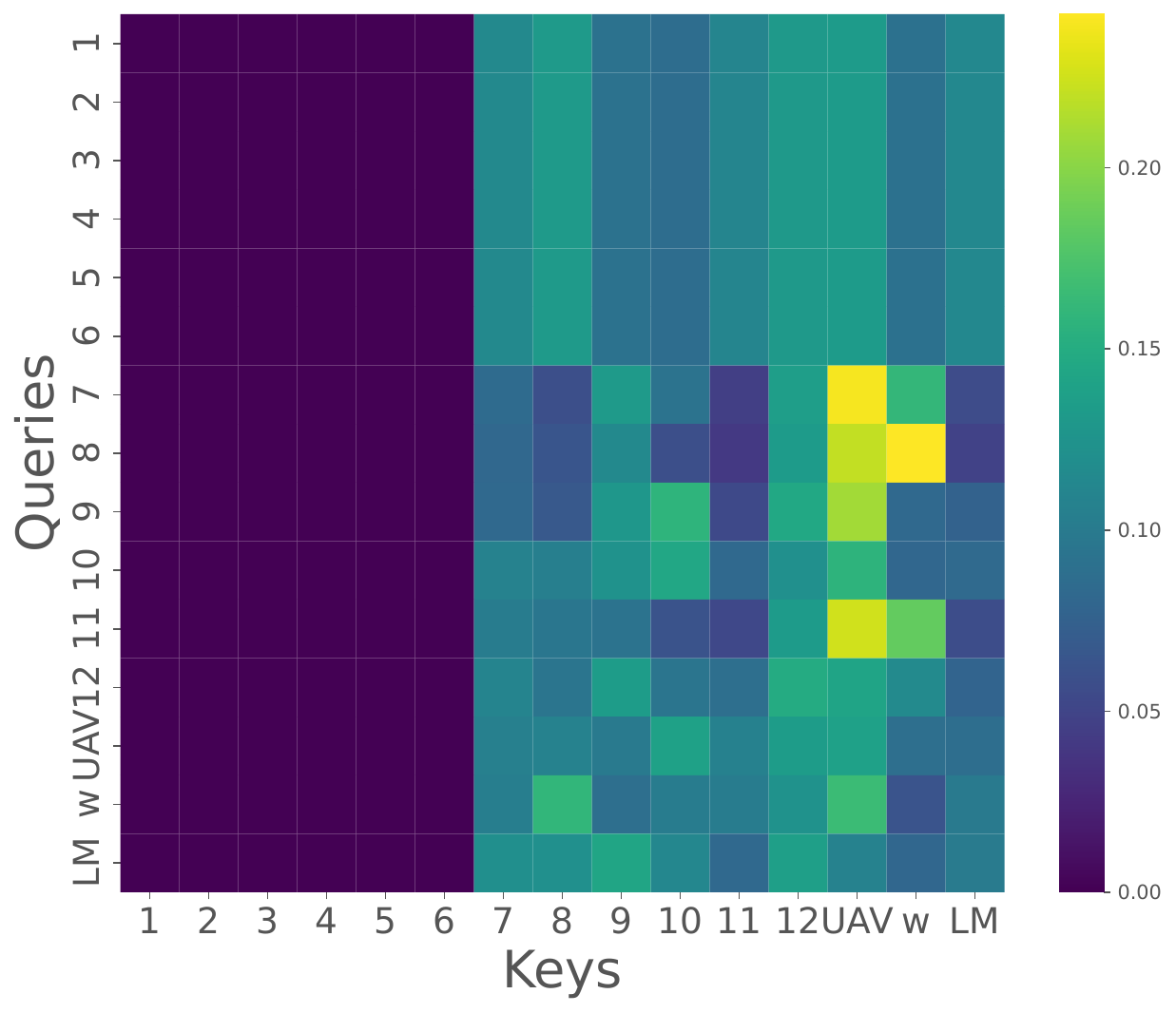}
    \caption{Matrix at initial state, $\boldsymbol{w} = (1,0)$}
    \label{fig:att_collection_1}
  \end{subfigure}%
  \hfill
  \begin{subfigure}{.32\textwidth}
    \centering
    \includegraphics[width=\textwidth]{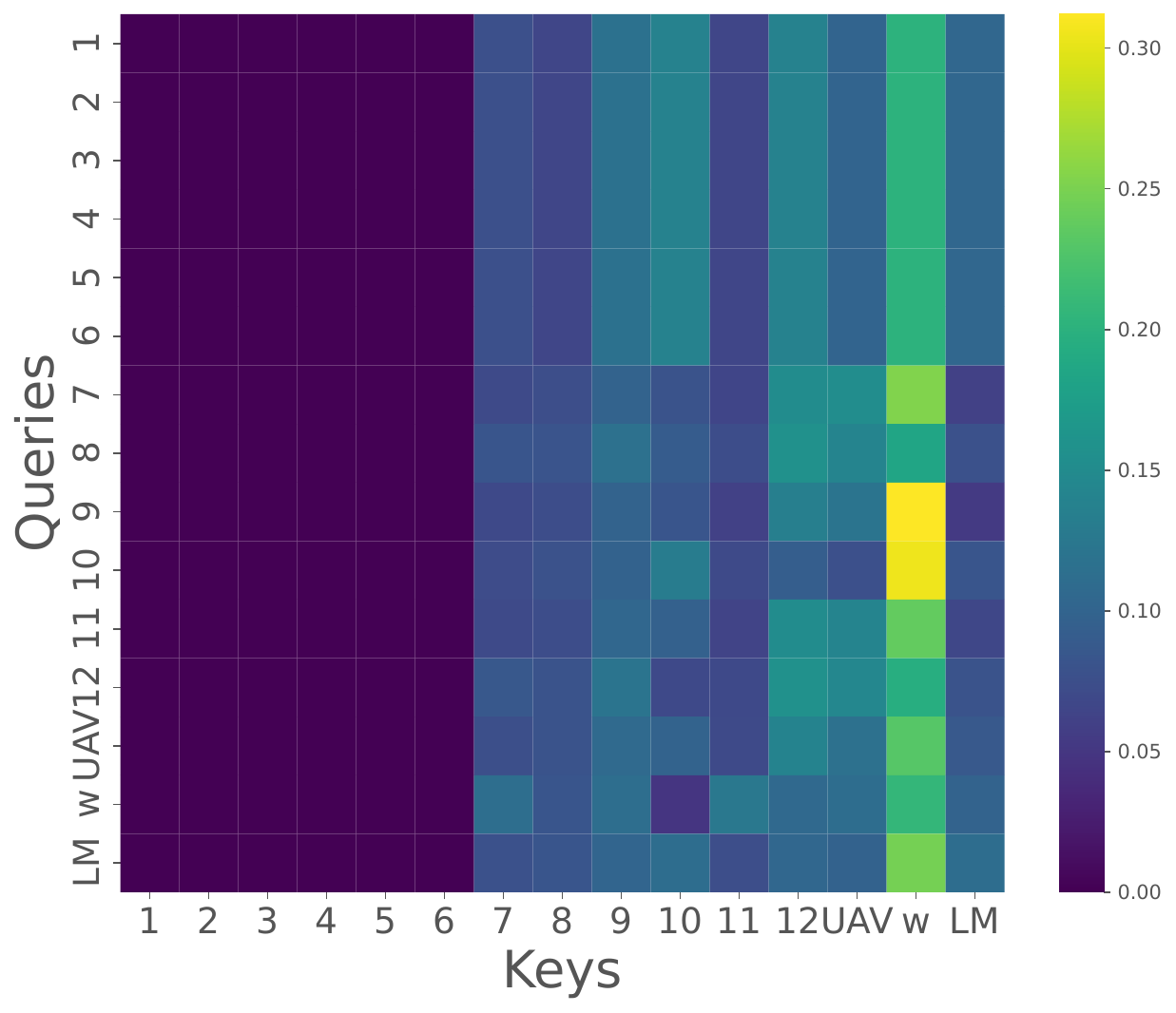}
    \caption{Matrix at initial state, $\boldsymbol{w} = (0,1)$}
    \label{fig:att_battery_1}
  \end{subfigure}%
  \hfill
  \begin{subfigure}{.32\textwidth}
    \centering
    \includegraphics[width=\textwidth]{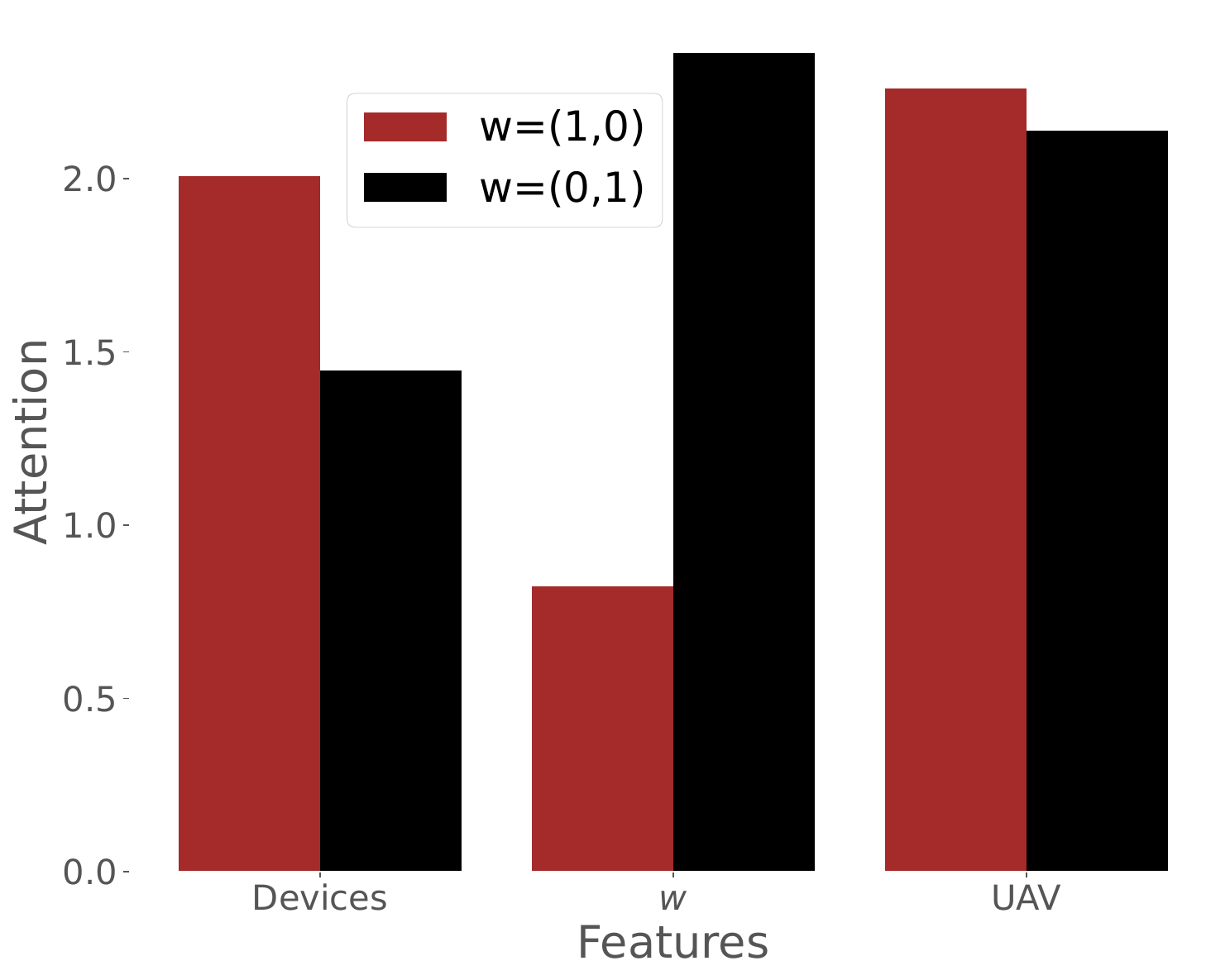}
    \caption{Averaged on complete trajectory}
    \label{fig:att_utility_1}
  \end{subfigure}%
  \hfill
  \caption{Attention scores in the RD scenario. (a) and (b) are examples of attention maps for the initial state of a realization with $6$ devices. (c) compares the average attention between the devices, the UAV, and the preference across the trajectory.
  }
  \label{fig:attention_map_1}
\end{figure*}

As previously discussed, we perform extensive evaluations in test settings that differ substantially from the training configuration. While the initial battery capacity is kept fixed at $b_0 = 80$ for most cases, the number of devices varies across $\{3, 5, 6, 9\}$. For the setting with $10$ devices, we increase the initial battery capacity to $b_0 = 100$ to account for the higher data collection demand. In Figure~\ref{fig:generalization_1}, we report three key metrics for each algorithm: (i) data collection performance under the preference $\boldsymbol{w} = (1, 0)$, (ii) battery consumption under $\boldsymbol{w} = (0, 1)$, and (iii) the average utility. The average utility for a given environment is defined as: $\text{Utility}= \frac{1}{|\mathcal{W}_{\text{test}}|}
   \sum_{\boldsymbol{w} \in \mathcal{W}_{\text{test}} }\boldsymbol{w}^T\boldsymbol{J}^{\pi}_w,$
with $\boldsymbol{J}^{\pi}_w$ denoting the multi-objective sum of rewards obtained following policy $\pi$ conditioned on $\boldsymbol{w}$.
Our proposed algorithm demonstrates superior generalization capabilities compared to the baselines, despite not relying on spatial map representations. Regarding the data collection objective, we observe that performance tends to improve when fewer than six devices are present. For instance, in the RD scenario, MOSAC-ATT is able to collect nearly all available data across all $100$ environment realizations when only three devices are present.
As the number of devices increases, the data collection rate naturally declines due to the greater demand on resources. However, MOSAC-ATT and MOSAC-CGL show notable improvements when the initial battery capacity is increased to $b_0 = 100$ in the $10$-device setting. In this case, MOSAC-ATT collects, on average, around $80\%$ of the data in both scenarios, outperforming all other methods. The next best performer, MOSAC-CGL, achieves an average of $72\%$.
In contrast, DDQN-CGL does not show any benefit from the increased battery capacity, highlighting its limited ability to adapt to more demanding conditions.
Additionally, thanks to our proposed architecture, MOSAC-ATT consistently outperforms the GREEDY baseline in terms of data collection across all tested settings and scenarios. In contrast, MOSAC-CGL and DDQN-CGL are outperformed by GREEDY in certain configurations, for example, in the RB scenario with $3$ devices and the RD scenario with $5$ devices, respectively.
This highlights a key distinction: GREEDY is inherently insensitive to variations in the environment, as it follows a deterministic policy that does not depend on data-driven training. In contrast, AI-based algorithms, including the RL approaches considered here, can be influenced by the distribution of the training data. The consistent performance of MOSAC-ATT in diverse and unseen environments underscores the strong generalization ability and robustness of our solution.
Generalization in terms of battery consumption is more straightforward to evaluate as the shortest path to the terminal zone is generally unaffected by changes in the number or positions of devices. As expected, the GREEDY algorithm performs optimally in all cases with respect to this objective.

In contrast, DDQN CGL and MOSAC CGL exhibit suboptimal behavior in several environment realizations, as evidenced by the discrepancy between their average battery consumption and the optimal values of $-32$ (RD) and $-2$ (RB), respectively, as well as by their non-zero standard deviations. Even in the RB scenario, where the optimal policy is trivial (e.g., taking a single step and then reversing it), MOSAC-CGL fails in some test environments with only $3$ devices.
Notably, the likelihood of such failures increases with the degree of mismatch between the test and training settings. This observation highlights a key limitation of RL-based methods: even when the optimal policy is simple and invariant for a given objective, agents can be misled by the variability and noise present in previously unseen test environments.
In contrast, MOSAC-ATT effectively mitigates this issue, thanks to its attention-based architecture. When the preference is set to $(0,1)$, the attention mechanism selectively focuses on features relevant to the battery consumption objective. This targeted focus enables the model to filter out irrelevant environmental variations and maintain optimal behavior across all settings and scenarios.
Figures~\ref{fig:gen_utility_1} and~\ref{fig:gen_utility_2} further confirm that the trends observed for edge-case preferences extend to the overall utility, strengthening the robustness of MOSAC-ATT.

\subsection{Attention Map Analysis}
To further support our interpretation of the generalization results, we analyze the attention scores produced by the second attention layer of the network. Figures~\ref{fig:att_collection_1} and~\ref{fig:att_battery_1} display the attention matrices corresponding to the initial state of a randomly selected environment realization in the RD scenario, under preferences $\boldsymbol{w}=(0,1)$ and $\boldsymbol{w}=(1,0)$, respectively. 

In each plot, the tokens, comprising the local map, preference vector, UAV information, and $K^{max}=12$ device embeddings, are represented along both the x-axis (keys) and y-axis (queries). The value at position $(q,\zeta)$ indicates how much the query token $q$ attends to the key token $\zeta$. To assess the total attention a feature receives, we compute the sum of its attention weights over all queries (i.e., vertically across each column).
Since the test environment contains only $6$ active devices, the remaining $6 = K^{max} - 6$ device tokens (columns) are masked and do not receive any attention.
The results reveal a clear shift in attention depending on the preference: when $\boldsymbol{w} = (1,0)$, the model primarily attends to the device tokens, while for $\boldsymbol{w} = (0,1)$, attention is focused almost entirely on the preference token. This behavior is further confirmed in Figure~\ref{fig:att_utility_1}, which shows the aggregated attention across a full episode, grouped by feature category (devices, UAV, and preference).
These findings demonstrate that the attention encoder dynamically modulates its focus based on the specified objective preference, effectively filtering out irrelevant map information. This selective mechanism contributes to the model's improved performance and generalization in unseen environments. We believe this property is particularly valuable in reinforcement learning, as it provides a high-level, interpretable mechanism by which the agent adapts its perception and decision-making according to the desired trade-off between objectives.

\section{Conclusions}
\label{sec:conclusion}

This paper proposed MOSAC-ATT, an attention-based MORL algorithm designed to address a UAV-enabled data collection problem in IoT networks. The task takes place in a realistic urban environment, where buildings obstruct communication links between the UAV and IoT devices.
In this setting, we tackled key challenges such as environmental variability and the optimization of the data collection–energy consumption trade-off, all without prior knowledge of user-defined objective preferences.
Leveraging a carefully designed state representation and the permutation invariance provided by the attention mechanism, MOSAC-ATT demonstrates significant improvements over spatial map-based RL methods in terms of sample efficiency, multi-objective performance, and generalization capability.
Furthermore, our approach offers promising levels of high-level interpretability and adaptability to unseen scenarios by dynamically attending to features most relevant to the current objective trade-off.
Future work may extend this framework to multi-UAV coordination, integrate partial observability, and model wireless interference, bringing the solution even closer to real-world deployment.

\bibliographystyle{IEEEtran}
\bibliography{ref}

\end{document}